\documentclass[runningheads]{llncs}

\usepackage{eccv}

\usepackage{eccvabbrv}

\usepackage{graphicx}
\usepackage{booktabs}

\usepackage[accsupp]{axessibility}  
\usepackage{mathtools,amsmath}
\usepackage{comment}

\usepackage{hyperref}

\usepackage{orcidlink}

\newcommand{\museNo}{01070\-421}

\newcommand{\DFVolNo}{PDC\-2022-133642-I00}

\begin{document}

\title{Multi-label out-of-distribution detection via evidential learning} 

\titlerunning{Multi-label out-of-distribution detection via evidential learning}

\author{Eduardo Aguilar \inst{1,3} 
\orcidlink{0000-0002-2463-0301} \and
Bogdan Raducanu \inst{2}\orcidlink{0000-0003-3648-8020} \and
Petia Radeva \inst{3}\orcidlink{0000-0003-0047-5172}}

\authorrunning{E. Aguilar et al.}

\institute{Dept. de Ingeniería de Sistemas y Computación, Universidad Católica del Norte, Antofagasta, Chile \\\email{eduardo.aguilar@ucn.cl} \and
Computer Vision Center, Universitat Autònoma de Barcelona, Barcelona, Spain \\
\email{bogdan@cvc.uab.es} \and
Dept. de Matemàtiques i Informàtica, Universitat de Barcelona, Barcelona, Spain\\
\email{petia.ivanova@ub.edu}}

\maketitle

\begin{abstract}

  A crucial requirement for machine learning algorithms is not only to perform well, but also to show robustness and adaptability when encountering novel scenarios. One way to achieve these characteristics is to endow the deep learning models with the ability to detect out-of-distribution (OOD) data, i.e. data that belong to distributions different from the one used during their training. It is even a more complicated situation, when these data usually are multi-label. In this paper, we propose an approach based on evidential deep learning in order to meet these challenges applied to visual recognition problems. More concretely, we designed a CNN architecture that uses a Beta Evidential Neural Network to compute both the likelihood and the predictive uncertainty of the samples. Based on these results, we propose afterwards two new uncertainty-based scores for OOD data detection: (i) OOD - score Max, based on the maximum evidence; and (ii) OOD score - Sum, which considers the evidence from all outputs. Extensive experiments have been carried out to validate the proposed approach using three widely-used datasets: PASCAL-VOC, MS-COCO and NUS-WIDE, demonstrating its outperformance over several State-of-the-Art methods.
  
\end{abstract}

\section{Introduction}
\label{sec:intro}

One of the main concerns of deploying deep learning systems to practical applications is represented by their capacity to prevent and avoid risks. In other words, AI systems need to show robustness against different types of threats, i.e. to maintain its performance when facing new, unrelated data. Of course, real-life scenarios always involve new or ambiguous scenarios, where the trustworthiness of the AI systems is challenged, because typically not all data are available from the beginning of the training stage. In consequence, changes in performance could manifest in two different manners, and some of
them may be unpredictable: either due to deliberate adversarial attacks, such as data poisoning, or as a consequence of natural occurrences, such as image corruption or unseen data. 

The focus of the current work is on the latter aspect, i.e. to increase the systems' robustness when facing with data not seen during training, also known as out-of-distribution (OOD) data \cite{hendrycks2017iclr}. A comprehensive survey on OOD detection can be found in \cite{yang2024surveyOOD} and recently a generalized benchmark has been proposed \cite{yang2022benchmarkOOD} in order to unify the evaluation methodologies of several existing methods. As a matter of fact, most of the existing work in OOD detection assumes 
that data is single-labeled
\cite{yang2024surveyOOD}. 
However, this could be an unrealistic scenario since in real-world applications data often present multiple labels, e.g. in scene understanding \cite{chen2019tip,zhu2023iccv}, object recognition \cite{cohen2021iccv}, semantic segmentation \cite{he2022eccv}, image retrieval \cite{li2023icmr}, etc. The OOD detection for multi-label data is a much more challenging scenario, because in this case the decision that a sample is OOD should be taken by considering the uncertainty of different labels, compared to the case when it corresponds to the most dominant label. 

For reliable detection of OOD data, it is very important to assess the uncertainty in the classification of deep learning algorithms.
Therefore, the field of uncertainty estimation in neural networks has seen a growing research interest \cite{abdar2021uncertaintysurvey}, including methods based on Bayesian Neural Networks \cite{maddox2019bayesian}, Deep Ensembles \cite{rahaman2021uncertainty} and Single Deterministic Neural Networks \cite{van2020uncertainty}. However, for the purpose of the current work, we will rely on Evidential Deep Learning (EDL) \cite{sensoy2018evidential} due to its ability to pinpoint various sources of uncertainty.  
Within this approach, the network's output is used to set the parameters of a Dirichlet distribution on the class probabilities. These parameters can then be used to assess the uncertainty of the network. The method has been successfully evaluated for the task of OOD detection~\cite{sensoy2018evidential} for the case of single-label data. 
Inspired by \cite{sensoy2018evidential}, in the current work, we propose an EDL-based deep learning approach for multi-label classification and evaluate the performance of several posthoc ODD data detection
methods on three common multi-label classification benchmarks: PASCAL-VOC, MS-COCO and
NUS-WIDE.

Regarding OOD detection, there are mainly two strategies, namely {\em semantic shift} and {\em distribution shift}. The semantic shift divides the dataset into two partitions: one partition is considered in-distribution, while the left-over partition, coming from the same distribution, is considered OOD. 
Alternatively, domain shift implies that one dataset is considered in-distribution data and other datasets,
which come from different distributions, 
represent OOD. 
In our study, we will address the OOD detection from the domain shift perspective.

Our main contributions are as follows:

\begin{itemize}
    \item We are the first to propose EDL for OOD detection in a multi-label classification setting.
    \item We propose two new uncertainty-based scores (using \textit{positive evidence}  and \textit{negative evidence}) for OOD data detection in the case of a multi-label classification setting.
    \item We extensively validate our approach on three common multi-label datasets, PASCAL-VOC, MS-COCO and NUS-WIDE, outperforming the existing state-of-the-art.
\end{itemize}

In the following section, we present a brief review of the works most related to our research. In Section~\ref{sec:method}, we describe our proposal for multi-label OOD detection. In Section~\ref{sec:experiments}, we describe the datasets, detail the experimental setup, and define the metrics for the validation. In Section~\ref{sec:results}, we show and discuss our results. Finally, we present conclusions and future directions.

\section{Related Work}
\label{sec:relatedwork}

We briefly introduce uncertainty estimation methods and review the most relevant work in OOD detection for single- and multi-label object recognition.

\subsubsection{Uncertainty Estimation.}
Uncertainty is primarily categorized into two types: epistemic uncertainty, which arises from insufficient training data, and aleatoric uncertainty, which is associated with random noise during sample collection \cite{kendall2017uncertainties}.
To estimate uncertainty, several deep neural methods have been developed, mainly for classification and regression problems \cite{gawlikowski2021survey}. These include Bayesian methods, which learn a distribution over the weights \cite{blundell2015weight};  ensemble methods, which are based on interpreting Bayesian methods as an ensemble of thin networks with shared weights \cite{lakshminarayanan2017simple}; test-time augmentation methods, which perform multiple forwards passes by randomly altering the input data to perform the prediction and the quantification of the uncertainty \cite{wang2019aleatoric}; and single deterministic methods, such as those based on evidence theory \cite{sensoy2018evidential}. In our work, we have been inspired by the latter due to the good results shown in OOD detection for single-label recognition \cite{sensoy2018evidential} and extend its approach to multi-label settings. 

\subsubsection{Single-Label OOD.}
In recent years, several OOD detection approaches have been proposed \cite{yang2024surveyOOD}, which can be divided into four categories: classification-based, density-based, distance-based and reconstruction-based methods. Some classification-based methods rely directly on the confidence of the classifier, i.e. samples with low likelihoods are considered OOD \cite{devries2018OOD, wang2021iccv}. Some other techniques focus on training strategies, such as logit normalization \cite{wei2022icml} or outlier synthesis \cite{du2022iclr,du2023neurips}. Density-based methods estimate the probability density of the training data and identify OOD samples as the ones that deviate from this distribution \cite{abati2019cvpr,xiao2020neurips,wang2022cvpr}. Compared to classification-based method, density-based methods are more difficult to use since they rely on generative models whose training and optimization are still challenging problems. Distance-based methods leverage the idea that OOD samples tend to lie farther from the in-distribution data in some feature space. These methods measure distances between test samples and in-distribution data points or centroids to determine if a sample is OOD. The existing approaches could be either parametric (based on Mahalanobis distance) \cite{lee2018neurips,ren2021arxiv} or non-parametric (based on nearest-neighbor distance) \cite{lee2018neurips,ren2021arxiv}. Finally, reconstruction-based methods rely on the premise that models trained on in-distribution data will reconstruct in-distribution samples accurately, but fail to do so for OOD samples. Existing approaches apply a reconstruction model to the whole image \cite{denouden2018arxiv}, masking a random part of it \cite{yang2022eccv,li2023cvpr} or to the hidden features \cite{zhoucvpr2022}.

\subsubsection{Multi-Label OOD.}
While OOD detection of single-label data clearly dominates the literature, recently there has been an increasing interest to extend the OOD detection to multi-label data. An energy-based approach has been proposed in \cite{wang2021neurips} which estimates the OOD data by aggregating label-wise energy scores
from multiple labels. The approach is based on an energy model that assigns a scalar ({\it energy score}) to an input sample: data belonging to the training set tend to have lower energy, while OOD data have higher energy. It was demonstrated theoretically and empirically that the energy score is superior to both softmax-based score and
generative-based methods for OOD detection. 
Another approach has been proposed in \cite{wang2022ivc} which exploits the co-occurrence and sparsity properties of labels for distinguishing the OOD data from in-distribution data. Their approach consists of three steps: (i) filtering out low-confidence predictions to simulate label sparsity; (ii) counting label co-occurrences from high-confidence predictions and the label co-occurrence matrix; and (iii) using these counts as weights to calculate the final OOD detection score.

Regarding practical, real-world applications, multi-label OOD data detection has been applied to image classification \cite{hendrycks2022icml}, action recognition \cite{zhao2023open}, and sewer defect classification \cite{zhao2023towards}.

\section{Methodology}
\label{sec:method}
In an open set scenario, it is desirable to provide an accurate and robust solution to address the classification of data that may or may not belong to the classes seen during model training. Uncertainty quantification in deep networks has proven to be effective in detecting data of unknown categories (e.g. OOD) while maintaining and even improving accuracy over traditional methods. 
The uncertainty-based methods have recently been used in related tasks, but have not been thoroughly compared for OOD data detection in the framework of a multi-label visual recognition problem (at least using the common benchmark 
represented by the three widely used datasets: PASCAL-VOC, MS-COCO and NUS-WIDE). 
In this paper, we propose an EDL-based deep learning approach to address the aforementioned challenges. 
Furthermore, we provide several uncertainty-based measures to explore its usability to determine when an input image corresponds to in-distribution or OOD data.  

\begin{figure}[!thb]
    \centering
    \includegraphics[width=\textwidth]{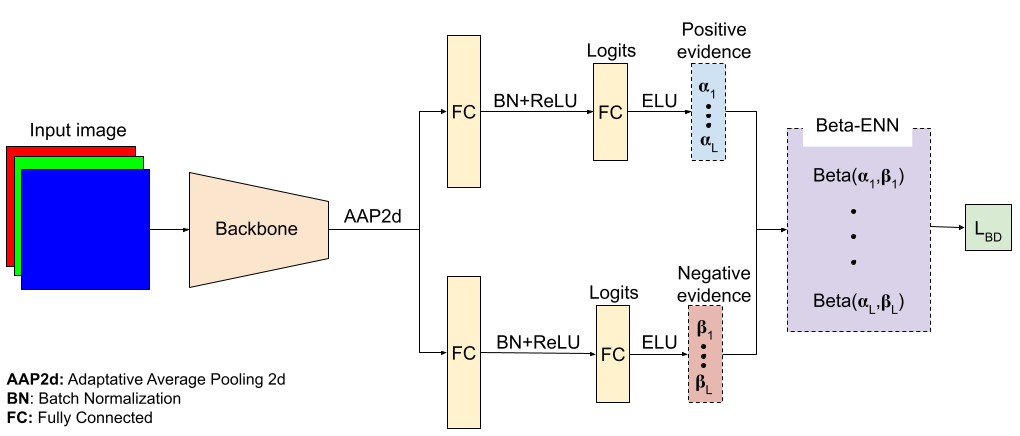}
    \caption{Overview of the Beta Evidential Neural Network proposed for multi-label recognition and OOD detection.}
    \label{fig:pipeline}
\end{figure}

\subsection{Multi-label Evidential Deep Learning}

Multi-label visual object recognition is a line of research that aims to classify an input image with multiple categories. Traditional approaches use a sigmoid activation after the logit layer to provide an independent likelihood for each output and based on a threshold (e.g. $\tau \geq 0.5$) decide which labels are predicted. These probabilities are calculated from what the model already knows, but say nothing about the unknown categories (e.g. OOD data).

Uncertainty quantification has proven useful for improving the understanding of model behavior and, in particular, for dealing with OOD data \cite{abdar2021uncertaintysurvey}. We are interested in the existing methods based on Evidential Learning, because of their ability to quantify uncertainty without affecting model performance or the computational resources required for training and use. EDL was initially published to address the problem of single-label object recognition \cite{sensoy2018evidential}. The method has been developed from the perspective of the Evidence Theory, in which the model prediction corresponds to the parameters of the prior Dirichlet distribution instead of the softmax probability. In this way, the model can learn a distribution over the possible softmax outputs, rather than the point estimate of a softmax output, and thus quantify the predictive uncertainty. 

More recently, an EDL-based approach has been proposed for multi-label open set action recognition \cite{zhao2023open}. Since in a multi-label problem, the likelihood of each label follows a binomial distribution, the conjugate prior is not a Dirichlet distribution (like in a single-label problem), but a Beta distribution. Motivated by this work, we design a CNN architecture that uses a Beta Evidential Neural Network (Beta-ENN) to compute both the likelihood and the predictive uncertainty (see Fig. \ref{fig:pipeline}). 
The Beta-ENN predicts subjective opinions instead of class probabilities, relying on the Evidence theory and Subjective Logic (SL) \cite{jsang2018subjective}. In multi-label settings, the subjective opinion $w_l = (b_l, d_l, u_l, a)$ for the l-th category is defined by the belief ($b_l$), disbelief ($d_l$), uncertainty mass ($u_l$), and base rate distribution ($a$), which represent prior knowledge. Here, $b_l$, $d_l$ and $u_l$ take values in the range $[0-1]$ and their sum equals to $1$. 
The binomial opinion follows a Beta probability density function, characterized by positive ($\alpha$) and negative ($\beta$) evidence. From this,
through SL, the principles of evidential theory are used to quantify the opinion $w_l$ in the proposed Beta-ENN method as follows: 
$$b_l = \frac{\alpha_l - aW}{\alpha_l + \beta_l} , \: \: d_l = \frac{\beta_l - aW}{\alpha_l + \beta_l} , \: \: u_l = \frac{W}{\alpha_l + \beta_l}, $$
where non-informative prior weight $W$ and base rate $a$ are empirically set to 0.5 and 2, respectively.

Specifically, given an input image $x$ and output logits $f^p$ and $f^n$, positive evidence ($\alpha = \alpha_1, \alpha_2, \ldots, \alpha_L$) and negative evidence ($\beta = \beta_1, \beta_2, \ldots, \beta_L $) are obtained by applying nonlinear activation on the output logits, in our case the activation is ELU. To ensure evidence greater than 1, we increase the output by 2 after applying ELU. Formally, $\alpha = ELU(f^p(x)) + 2$ and $\beta = ELU(f^n(x)) + 2$. With the parameters of the Beta distribution, the prediction of the model for the l-th category, which represents the mean of the distribution, is given by:
$$p_l = \frac{\alpha_l}{\alpha_l + \beta_l}.$$

Let $y$ be a one-hot vector encoding the ground-truth labels of x, in a Beta-ENN the learning of subjective opinions can be realized by a Beta loss function ($L_{BD}$) calculating its Bayes risk as follows:
\begin{equation}
\begin{split}
    L_{BD}(\theta) & = \frac{1}{L}\sum_{l=1}^{L}{\int \textbf{BCE}(y_l, p_l)\textbf{Beta}(p_l; \alpha_l, \beta_l)dp_l } \\
    & = \frac{1}{L} \sum_{l=1}^{L} [ y_l(\digamma(\alpha_l +\beta_l)-(\digamma(\alpha_l)) \\
    & + (1 - y_l)((\digamma(\alpha_l +\beta_l)-(\digamma(\beta_l))],
    \end{split}
\end{equation}

$$ \textbf{Beta}(p_l; \alpha_l, \beta_l) = \frac{p_l^{\alpha_l -1}(1-p_l)^{\beta_l -1}}{\textbf{B}(\alpha_l, \beta_l)},$$

$$ \textbf{B}(\alpha_l, \beta_l) = \frac{\Gamma(\alpha)\Gamma(\beta)}{\Gamma(\alpha)+\Gamma(\beta)},$$

$$ \textbf{BCE}(y_l, p_l) = y_l \text{log}(p_l) + (1 - y_l)\text{log}(1 - p_l),$$

where $\textbf{BCE(.,.)}$ is the Binary Cross Entropy loss, $\textbf{Beta(.,.,.)}$ is the PDF of the prior Beta distribution for drawing the class probability for the input sample, \textbf{B} is the Beta function, $\Gamma(.)$ is the gamma function and $\digamma(.)$ is the digamma function. 

\subsection{Multi-label EDL-based OOD Scores}

As demonstrated in previous studies \cite{gawlikowski2023survey}, uncertainty plays an important role in detecting OOD data. In a Beta-ENN, uncertainty can be analyzed from the perspective of positive evidence, negative evidence, or both. 
Specifically, in a multi-label problem, we expect the input image to have at least one object that belongs to the categories learned by the model. Therefore, assuming the input data is in-distribution, we would expect the positive evidence for one of the labels to be high and the negative evidence for the same label to be low, resulting in a low $u$. However, although we obtain a low $u_l$ when $\alpha_l$ is high and $\beta_l$ is low, this also occurs when $\beta_l$ is high and $\alpha_l$ is low or when both are high. In the latter two cases, we cannot be sure that at least one object has been correctly predicted. For this reason, we propose uncertainty-based scores that are best suited to the problem at hand, analyzing the evidence separately or together to determine whether the data are in- or out- of the distribution.  We grouped the scores into two: \textit{OOD Score - Max}, which is based on the maximum evidence, and \textit{OOD Score - Sum}, which considers the evidence from all outputs. 

\subsubsection{OOD Score - Max:} As most posthoc OOD detection methods in single-label recognition, the score can be calculated taking into account the maximum-
valued statistics among all labels. For that, we propose the following three scores, which considered:
\begin {itemize}
\item the positive evidence ($U^m_{p}$): 
\begin{equation*}
     U^m_{p} = \frac{1}{\text{max}(\alpha)}
\end{equation*}
\item the negative evidence ($U^m_{n}$):
\begin{equation*} 
U^m_{n} = 1 - \text{max}(\beta^{-1}) 
\end{equation*} 
\item the positive and negative evidence ($U^m_{p,n}$): 
\begin{equation*} 
U^m_{p,n} = \lambda_1 \times U^m_{p} + (1-\lambda_1) \times U^m_{n} 
\end{equation*}  
\end{itemize}
where $U_p^m$ and $U_n^m$ are uncertainty-based scores that use the maximum positive and negative evidence among all labels, respectively. $U_{p,n}^m$ is a score that weights the contribution of the proposed $U_p^m$ and  $U_n^m$ scores. 
Note that in $U_p^m$, the lower the positive evidence, the higher the probability that the target data is OOD. In $U_n^m$, the higher evidence of the label with lower negative evidence, the higher the probability that the target data is OOD. 

\subsubsection{OOD Score - Sum:} In multi-label OOD detection, as demonstrated in \cite{wang2021neurips}, it is more appropriate to calculate the score taking into account all the outputs. For that, we propose the following three scores, which consider:
\begin {itemize}
\item the positive evidence ($U^s_{p}$):
\begin{equation*} 
U^s_{p} = \frac{L}{\sum_{l=1}^{L}{\alpha_i}},
\end{equation*} 
\item the negative evidence ($U^s_{n}$): 
\begin{equation*}
U^s_{n} = 1 - \frac{\sum_{l=1}^{L}{\beta_i^{-1}}}{L},
\end{equation*}
\item the positive and negative evidences ($U^s_{p,n}$):  
\begin{equation*}
U^s_{p,n} = \lambda_2 \times U^s_{p} + (1-\lambda_2) \times U^s_{n}, 
\end{equation*} 
\end{itemize}
where $U_p^s$ and $U_n^s$ are uncertainty-based scores that consider the sum of the positive and the inverse of the negative evidence for all labels, respectively. $U_{p,n}^s$ is a score that weights the contribution of the proposed $U_p^s$ and  $U_n^s$ scores. 
Note that in $U_p^s$, the lower the total positive evidence, the higher the probability that the target data is OOD. In $U_n^s$, the higher the total of the inverse of the negative evidences, the higher the probability that the target data is OOD. 

\section{Experiments}
\label{sec:experiments}

In this section, we describe the datasets used, detail the implementation results and define how the validation of the proposed method is performed.

\subsection{Datasets}

Similar to previous works \cite{wang2021neurips, wang2022ivc}, we use three widely used datasets PASCAL-VOC \cite{everingham2015ijcv}, MS-COCO \cite{lin2014eccv} and NUS-WIDE \cite{chua2009civr} to evaluate our model's performance in detecting OOD data.

\subsubsection{PASCAL-VOC} (2012) contains a large collection of images (22,531) across
20 classes, featuring everyday objects and scenes (cars, people, animals, etc). Each image comes with bounding boxes and labels for a specific set of object classes. The images are distributed as 5,717 for training, 5,823 for validation and 10,991 for testing.

\subsubsection{MS-COCO} (2014) consists of 82,783 training, 40,504 validation, and 40,775 testing images with 80 common object categories (people, animals, vehicles, and household items, etc.). The images depict complex everyday scenes with common objects in their natural contexts, typically with multiple objects per image.  The annotations are detailed, providing not only class labels, but also instance segmentation and keypoints for human body parts.

\subsubsection{NUS-WIDE}
includes 269,648 images collected from Flickr across 81 concept labels, ranging from everyday objects like 'cars' to scenes like 'birthday party' or even emotions like 'happiness'. Each image is annotated with multiple labels (concepts) to reflect the diverse content within the image. Since NUS-WIDE has invalid and untagged images, we follow \cite{zhu2017learning} and use 119,986 training images and 83,389 test images. 
The test set is divided into validation and test sets with 41,949 data each. 

\subsection{Implementation}

We follow the protocol for OOD detection evaluation used in \cite{wang2021neurips} to compare the proposed approach with existing posthoc methods such as: MaxLogit \cite{hendrycks2022icml}, MSP \cite{hendrycks2016baseline}, ODIN \cite{liang2018enhancing}, Mahalanobis (M) \cite{lee2018neurips}, LOF \cite{breunig2000icmd}, Isolation Forest (ISOF) \cite{liu2008icdm}, and JointEnergy \cite{wang2021neurips}. 
For all experiments, ResNet101 \cite{he2016deep} pre-trained on ImageNet-1K \cite{deng2009cvpr} is used as the backbone and trained for $20$ epochs with a batch size of $64$. Optimization is performed using SGD with an initial learning rate of $1e-5$ for the backbone and $1e-4$ for the head. 

\begin{figure}[h]
    \centering
    \includegraphics[width=0.5\textwidth]{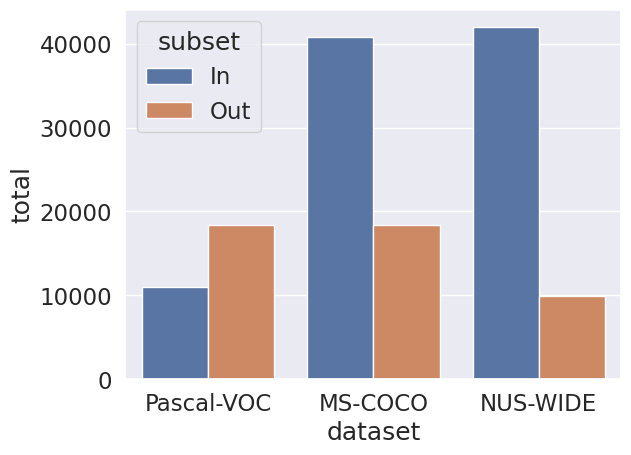}
    \caption{Total of in- and out- of distribution instances used to evaluate the performance on each dataset.}
    \label{fig:instance_distribution}
\end{figure}

The methods are evaluated on PASCAL-VOC, MS-COCO and NUS-WIDE, three widely used datasets for multi-label classification. 
The test set of each dataset was considered as in-distribution data and the data from a subset of ImageNet-22K as out-of-distribution data. The data from ImageNet-22K was selected in the same way that \cite{wang2021neurips}. The number of samples from each subset used for OOD detection for each dataset can be seen in Fig. \ref{fig:instance_distribution}.

As for data processing, the images were normalized by scaling the pixels from 0 to 1, subtracting the mean, and dividing the standard deviation. The mean and standard deviation were calculated from the ImageNet-1K images. During training, traditional data augmentation strategies were applied: random horizontal flipping, random scaling (from 0.5 to 2.0) and random crops with a size of $256 \times 256$. During validation, the images were resized to $256 \times 256$.

With respect to the proposed OOD scores, the weighting parameters in $U_{p,n}^m$ and $U_{p,s}^m$ were set considering an equal contribution of positive and negative evidence-based scores:  
$\lambda_1 = 0.5$ and $\lambda_2 = 0.5$.

All experiments were performed using PyTorch library on a computer with a NVIDIA RTX 2080 TI graphics card.

\subsection{Metrics for Evaluation}

OOD detection performance is calculated as a binary classification problem using the area under the receiver operating characteristic curve (AUROC), the area under the accuracy-recall curve (AUPR) and the false positive rate at the $95\%$ of recall (FPR95). A higher value of AUROC and AUPR, and a lower value of FPR95, indicate better detection performance.

\section{Results}
\label{sec:results}

To compare the performance of the different OOD scores, the same method was trained 5 times with random initialization of weights. The results of the multi-label recognition in terms of mean Average Precision (mAP) are shown in Table\;\ref{tab:ml_results}. After training, the OOD scores were applied to each method to determine the OOD data. The mean and median OOD detection results are presented in the tables and used in the graphs, respectively.

\begin{table}[]
    \caption{Multi-label object recognition performance on the validation set of PASCAL-VOC, MS-COCO and NUS-WIDE.}
    \centering
    \begin{tabular}{l|c|c|c|c|c}
        \hline
         \multicolumn{1}{c}{}& \multicolumn{5}{c}{mAP}  \\
         \multicolumn{1}{c}{Datasets} & \multicolumn{1}{c}{Beta-ENN1} & \multicolumn{1}{c}{Beta-ENN2} & \multicolumn{1}{c}{Beta-ENN3} & \multicolumn{1}{c}{Beta-ENN4} & \multicolumn{1}{c}{Beta-ENN5}  \\
         \hline
         PASCAL-VOC & 88.14 & 87.86 & 87.48 & 87.57 & 88.11 
         \\
         MS-COCO & 76.61 & 75.65 & 76.19 & 76.36 & 77.11
         \\
         NUS-WIDE & 60.48 & 59.80 & 60.71 & 58.38 & 59.79 \\
         \hline
    \end{tabular}
    \label{tab:ml_results}
\end{table}

Table \ref{tab:unc_based_score} shows the results with the different OOD scores proposed in the framework of a Beta-ENN. The first three scores correspond to the maximum criterion and the last three to the sum criterion. As can be seen, scoring based on positive evidence tends to perform much better than those based on negative evidence. Likewise, it is observed that, regardless of the aggregation criteria, the positive and negative evidence-based scores recover complementary information that allows for improving the results when both are considered together. 
The only exception occurs in the NUS-WIDE dataset for Max aggregation, where $U_p^m$ is better than $U_{p,n}^s$. We assume that, due to the complexity of the dataset itself (showing the worst performance in Table \ref{tab:ml_results}), the model was not able to learn the negative evidence well, which adversely affects the performance when we integrate both evidences into the $U_{p,n}^m$ score. 
Comparing the best methods for each aggregation criterion shows that the sum criterion greatly outperforms compared to the maximum criterion, highlighting the importance of the outputs of all labels for detecting OOD data. As can be seen, the best proposed OOD score is $U_{p,n}^s$, which is used for the analysis of the rest of the results. 

\begin{table}[!h]
    \caption{OOD detection performance for each proposed OOD score. The first three rows correspond to the Max criterion and the last three to the Sum criterion.}
    \centering
    \begin{tabular}{l|c|c|c|c|c|c|c|c|c}
         \hline
         \multicolumn{1}{l}{$\mathcal{D}_{in}$} & \multicolumn{3}{c}{PASCAL-VOC}& \multicolumn{3}{c}{MS-COCO} & \multicolumn{3}{c}{NUS-WIDE}\\
         \multicolumn{4}{c}{}& \multicolumn{1}{c}{FPR95} & \multicolumn{1}{c}{AUROC} & \multicolumn{1}{c}{AUPR}& \multicolumn{3}{c}{} \\
         \multicolumn{1}{l}{\textbf{OOD Score}} & \multicolumn{3}{c}{} & \multicolumn{1}{c}{$\downarrow$} & \multicolumn{1}{c}{$\uparrow$} & \multicolumn{1}{c}{$\uparrow$} &
         \multicolumn{3}{c}{}\\
         \hline
         $U_p^m$ & 48.30  & 87.98  & 79.40  & 49.27  & 87.21  & 93.11  & \textbf{52.07}  & \textbf{87.06}  & \textbf{95.97}  \\
         $U_n^m$ & 50.46  & 81.06  & 63.30  & 46.31  & 81.07  & 86.23  & 87.58  & 62.27  & 85.52   \\
         $U_{p,n}^m$ & \textbf{47.74} & \textbf{88.16} & \textbf{79.61} & \textbf{44.85} & \textbf{88.12} & \textbf{93.52} & 70.31 & 69.72 & 90.14 \\
         \hline
         $U_p^s$ & 41.69 & 82.76 & 85.97 & 38.84 & 91.43 & 95.59 & 46.34 & 88.20 & 96.35 \\
         $U_n^s$ & 52.27 & 80.97 & 70.92 & 49.62 & 84.54 & 91.97 & 96.50 & 53.27 & 83.08 \\
         $U_{p,n}^s$ & \textbf{40.30} & \textbf{91.28} & \textbf{86.54} & \textbf{37.20} & \textbf{91.94} & \textbf{95.95} & \textbf{46.18} & \textbf{88.58} & \textbf{96.51} \\
         \hline
          
    \end{tabular}
    \label{tab:unc_based_score}
\end{table}

Table \ref{tab:tab_1_sota} presents the OOD data detection results of several state-of-the-art (SoTA) methods. Most of them use the maximum score as a criterion, such as: MaxLogit, MSP, ODIN and Mahalanobis. Other approaches are based on local density (LOF) and tree (ISOF). In addition, more appropriate for the Multi-label problem, the sum criterion across all labels in JointEnergy and the proposed EDL-based metric $U_{p,n}^s$ are used. For all the metrics, except for $U_{p,n}^s$, it was used the logit layer placed before the positive evidence. The same method was trained five times using random initialization of the weights and then was used to evaluate the OOD detection performance. 
Interestingly, although it is a very simple criterion, MaxLogit offers the best performance compared to the other max-scoring criteria for the task at hand. Looking at the JointEnergy sum-based score, a vast improvement over MaxLogit is observed reinforcing the idea that the sum criterion is more suitable to perform OOD detection on multi-label data. The proposed $U_{p,n}^s$ criterion can further improve the results on all datasets, highlighting the advantages of uncertainty-based measures for this problem.

\begin{table}[!h]
    \caption{Benchmark evaluation of the SoTA and the proposed uncertainty-based score for OOD detection. }
    \centering
    \addtolength{\tabcolsep}{2pt}
    \begin{tabular}{l|c|c|c|c|c|c|c|c|c}
         \hline
         \multicolumn{1}{l}{$\mathcal{D}_{in}$} & \multicolumn{3}{c}{PASCAL-VOC}& \multicolumn{3}{c}{MS-COCO} & \multicolumn{3}{c}{NUS-WIDE}\\
         \multicolumn{4}{c}{}& \multicolumn{1}{c}{FPR95} & \multicolumn{1}{c}{AUROC} & \multicolumn{1}{c}{AUPR}& \multicolumn{3}{c}{} \\
         \multicolumn{1}{l}{\textbf{OOD Score}} & \multicolumn{3}{c}{} & \multicolumn{1}{c}{$\downarrow$} & \multicolumn{1}{c}{$\uparrow$} & \multicolumn{1}{c}{$\uparrow$} &
         \multicolumn{3}{c}{}\\
         \hline
         MaxLogit & 48.30 & 88.07 & 79.40 & 49.27 & 87.21 & 93.11 & 52.07  & 87.06 & 95.97 \\
         MSP & 82.89 & 67.27 & 49.50 & 86.82 & 64.67 & 77.09 & 86.38 & 63.22 & 85.63 \\
         ODIN & 51.56 & 87.44 & 77.99 & 50.06 & 85.72& 90.63 & 52.07 & 80.96 & 92.26 \\
         Mahalanobis & 52.24 & 84.29 & 70.32 & 71.26 & 78.49 & 88.25 & 82.94 & 73.76 & 92.10 \\
         LOF & 91.87& 63.50 & 51.50 & 94.71 & 62.27 & 79.73 & 97.82 & 49.68 & 82.32 \\
         Isolation Forest & 99.48 & 34.46 & 30.23 & 99.52 & 28.01 & 57.24 & 99.60 & 52.24 & 84.21 \\
         JointEnergy & 41.83 & 90.94 & 85.97 & 38.81 & 91.45 & 95.61 & 46.32 & 88.26 & 96.33 \\
         \hline
         EDL ($U_{p,n}^s$) & \textbf{40.30} & \textbf{91.28} & \textbf{86.54} & \textbf{37.20} & \textbf{91.94} & \textbf{95.95} & \textbf{46.18} & \textbf{88.58} & \textbf{96.51} \\
         \hline
          
    \end{tabular}
    \label{tab:tab_1_sota}
\end{table}

\begin{figure*}[!t]
        \centering
        \begin{subfigure}[b]{0.49\textwidth}
            \centering
            \includegraphics[width=\textwidth]{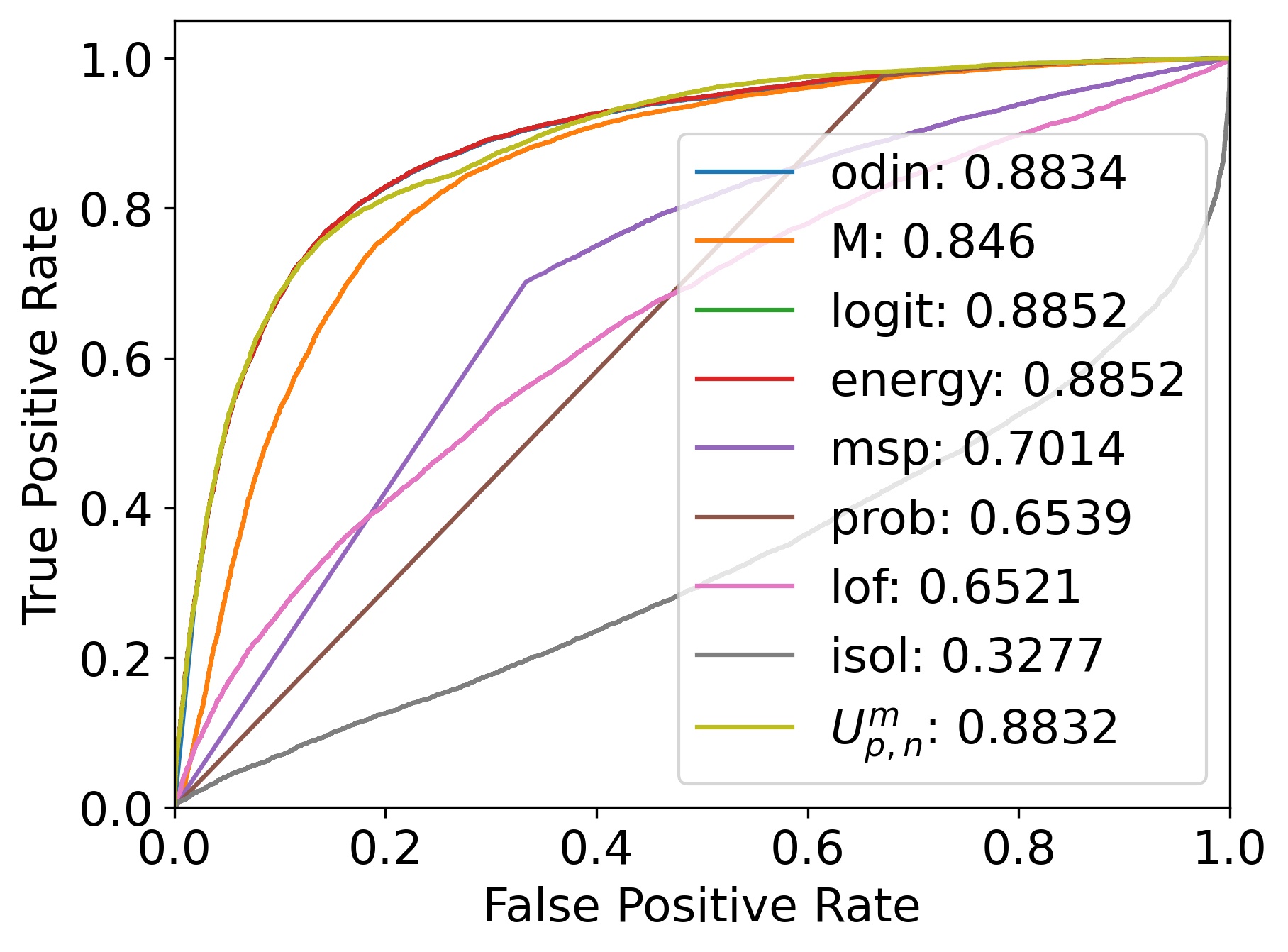}  
        \end{subfigure}
        \begin{subfigure}[b]{0.49\textwidth}  
            \centering 
            \includegraphics[width=\textwidth]{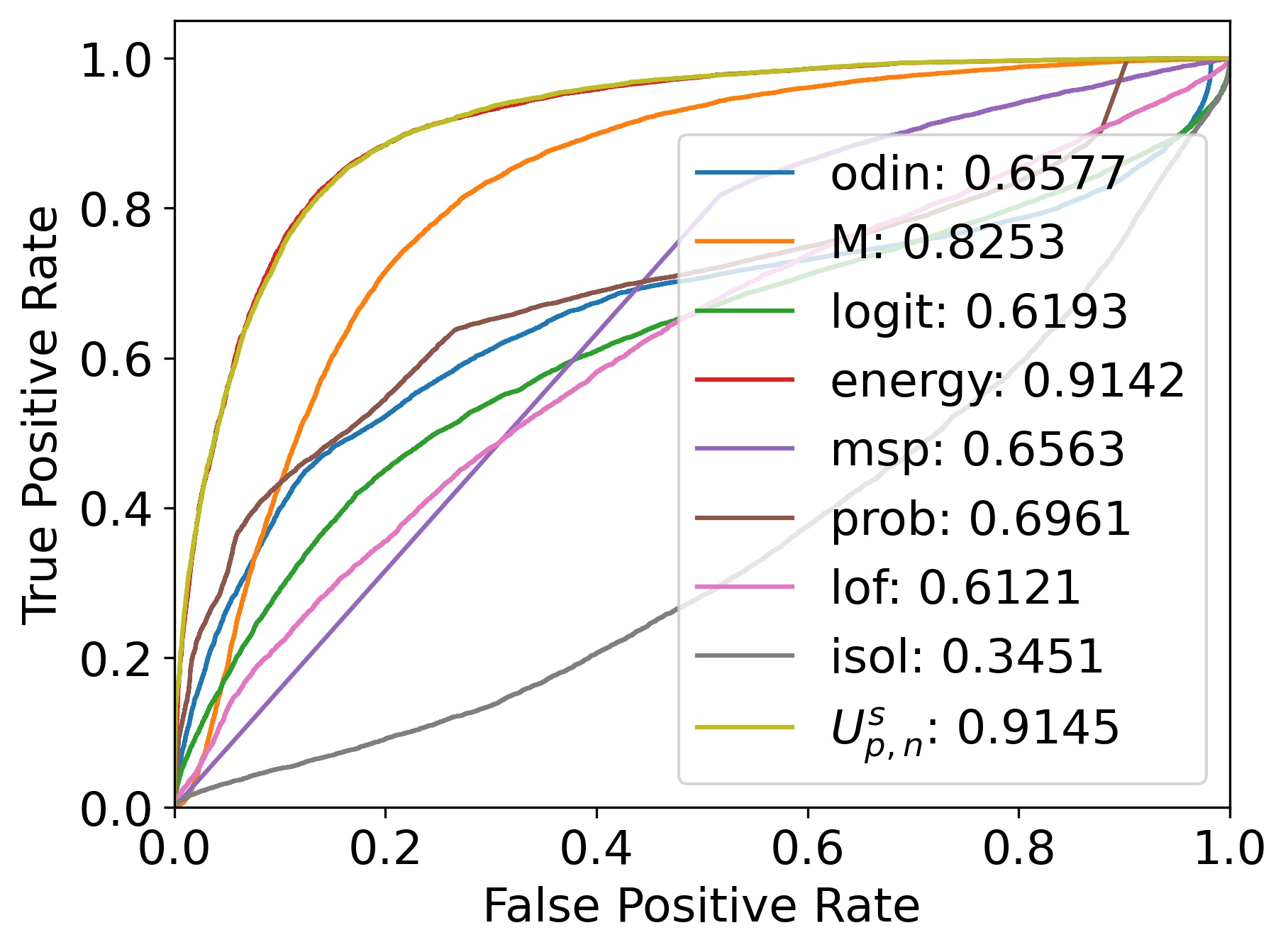}  
        \end{subfigure}
        \\
        \begin{subfigure}[b]{0.49\textwidth}  
            \centering 
            \includegraphics[width=\textwidth]{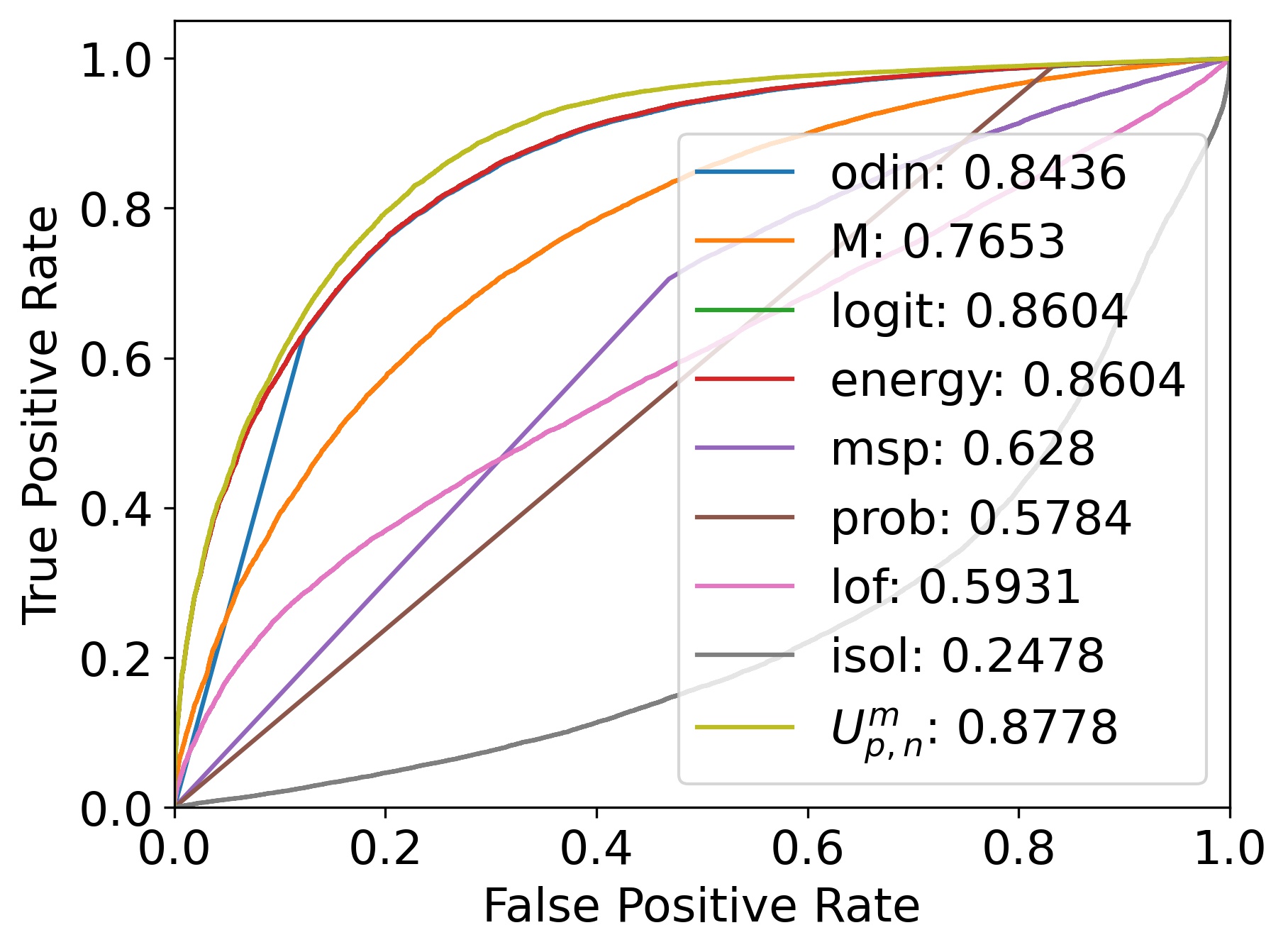}  
        \end{subfigure}
        \begin{subfigure}[b]{0.49\textwidth}
            \centering
            \includegraphics[width=\textwidth]{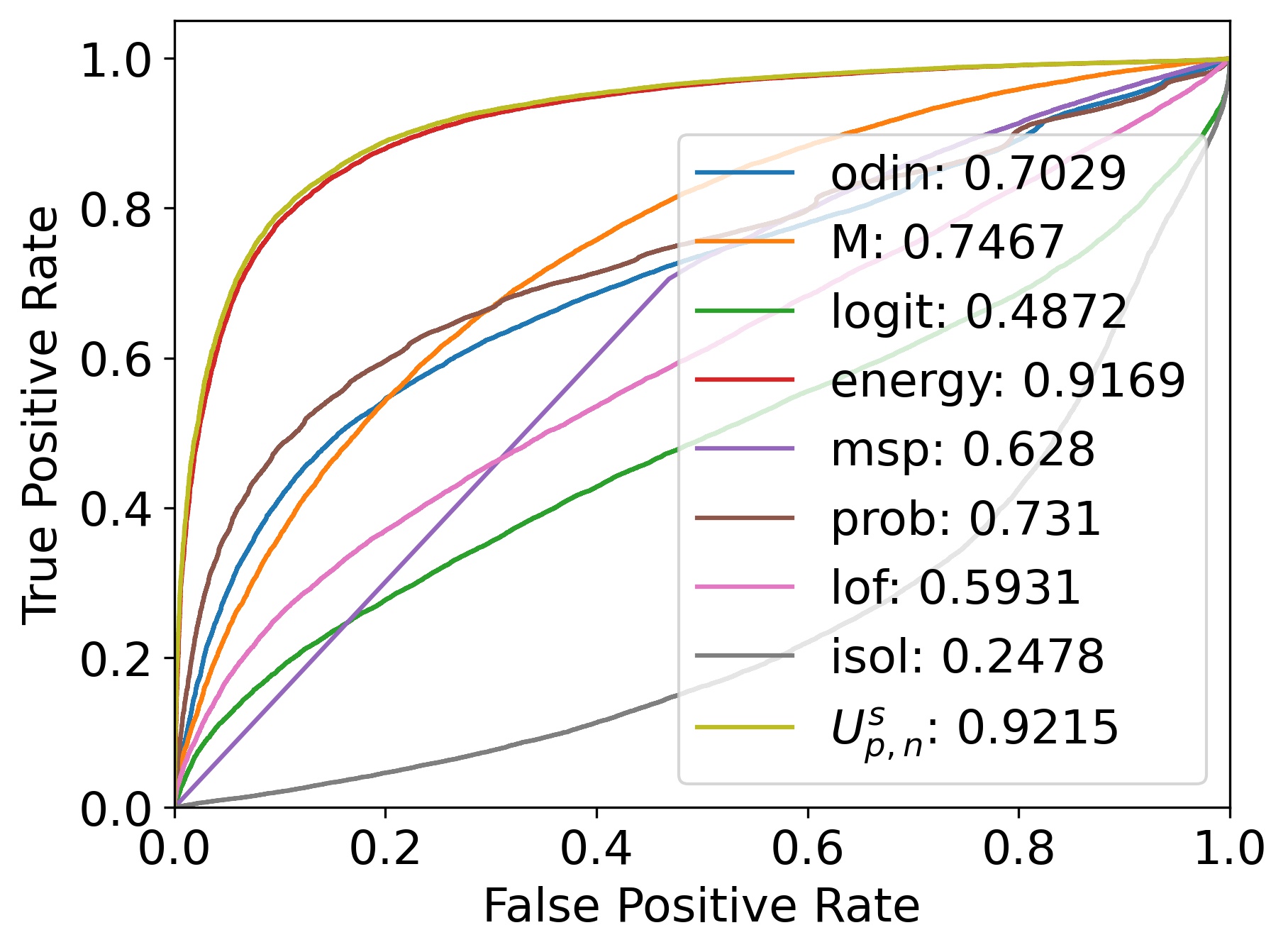} 
        \end{subfigure}
        \\
        \begin{subfigure}[b]{0.49\textwidth}  
            \centering 
            \includegraphics[width=\textwidth]{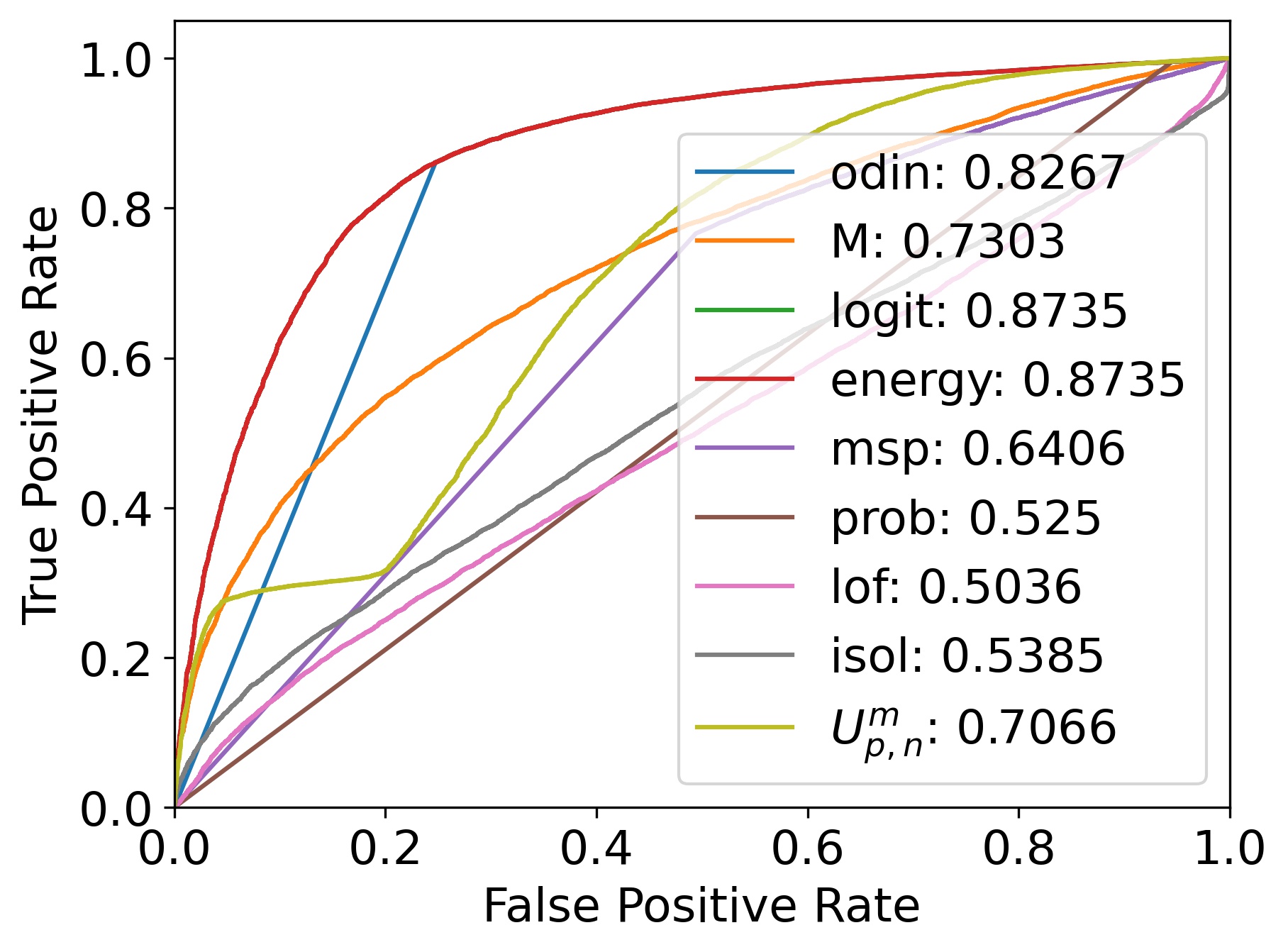} 
        \end{subfigure}
        \begin{subfigure}[b]{0.49\textwidth}  
            \centering 
            \includegraphics[width=\textwidth]{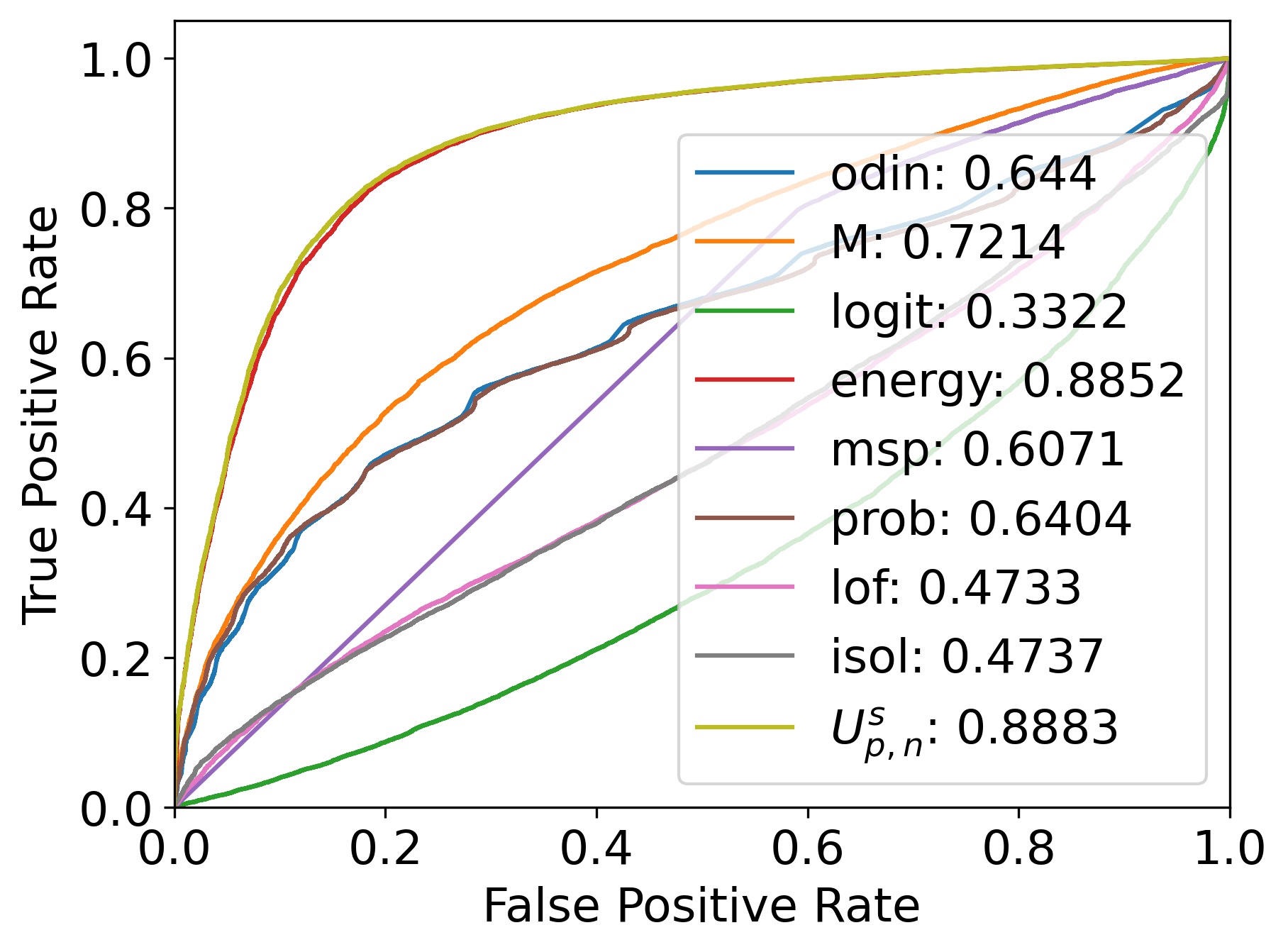} 
        \end{subfigure}
        \caption[  ]
        {\small ROC curve for OOD data detection. The first, second, and third rows correspond to the OOD performance on PASCAL-VOC, MS-COCO and NUS-WIDE, respectively. For each OOD score, the results are presented when using the Max (left column) or Sum (right column) criteria. } 
        \label{fig:roc}
    \end{figure*}

The ROC curve for all the OOD scores using the max and sum criterion is presented in Fig. \ref{fig:roc}. As can be seen, most of the OOD scores based on the maximum score reduce their performance when the criterion is replaced by the sum except for prob. This shows that this score is not appropriate for a multi-label problem where the result of all labels must be taken into account. Moreover, we can easily see that two methods are better than the rest, the energy method and the $U_{p,n}^s$, both using the sum criterion. The latter, $U_{p,n}^s$, has a slightly better performance in terms of AUROC. 

\begin{figure*} [!t]
        \centering
         \begin{subfigure}[b]{0.32\textwidth}
            \centering
            \includegraphics[width=\textwidth]{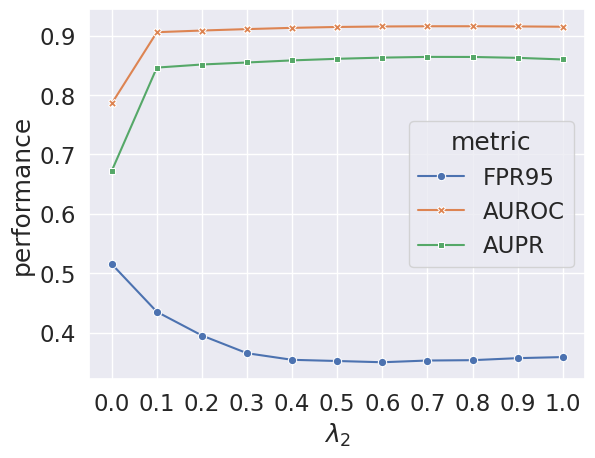}   
        \end{subfigure}
        \hfill
        \begin{subfigure}[b]{0.32\textwidth}  
            \centering 
            \includegraphics[width=\textwidth]{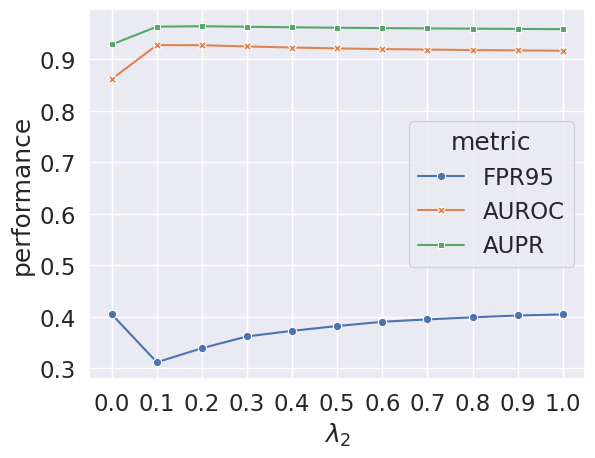}  
        \end{subfigure}
        \hfill
        \begin{subfigure}[b]{0.32\textwidth}  
            \centering 
            \includegraphics[width=\textwidth]{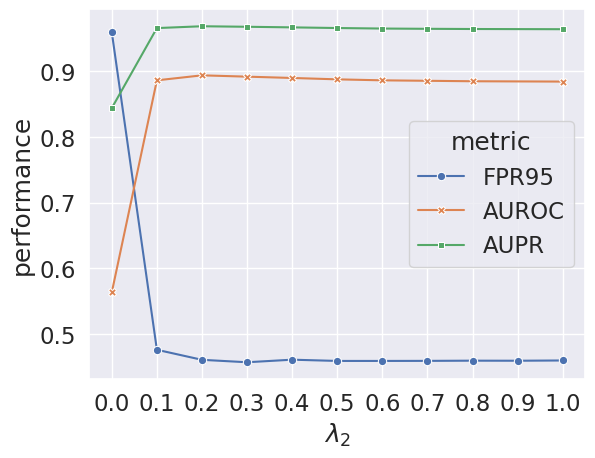}
        \end{subfigure}
        \caption[  ]
        {\small OOD detection performance by varying the value of $\lambda_2$ in the proposed $U_{p,n}^s$ approach. From left to right: PASCAL-VOC, MS-COCO and NUS-WIDE. } 
        \label{fig:lambda}
    \end{figure*}

Fig. \ref{fig:lambda} shows the OOD detection performance using the proposed $U_{p,n}^s$ when different values of $\lambda_2$ are used. When $\lambda_2$ is 0 it corresponds to $U_{n}^s$, when it is 1 it corresponds to $U_{p}^s$. The higher the values taken by $\lambda_2$, the greater the contribution of $U_{p}^s$ to the OOD score. Conversely, the lower the values the greater the contribution $U_{n}^s$. Different behaviors are observed for each data set, being more noticeable for FPR95 metric. In PASCAL-VOC, it is observed that the performance improves smoothly while increasing the contribution of the $U_{p}^s$ for the OOD score up to 0.7 and then decreases slightly. For MS-COCO, the best performance is obtained when a higher contribution of the $U_{n}^s$ is considered, specifically with $\lambda_2= 0.1$. And for NUS-WIDE, although the $U_{n}^s$ provide very poor performance when $U_{p}$ is used together, it is possible to improve the $U_{p}$ results a bit more with the proposed $U_{p,n}$. This demonstrates the benefits of combining both scores and the robustness of this metric for different $\lambda_2$ values. In addition, it is interesting to note that if the results are presented with the best $\lambda_2$, the improvement over JointEnergy (second best method) would be even greater. 

\begin{figure*}[!t]
        \centering
        \begin{subfigure}[b]{0.20\textwidth}
            \centering
            \includegraphics[width=\textwidth]{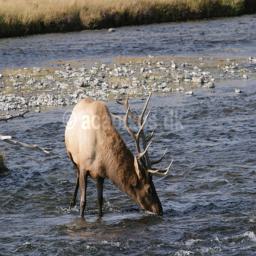} 
        \end{subfigure}
        \hfill
        \begin{subfigure}[b]{0.28\textwidth}  
            \centering 
            \includegraphics[width=\textwidth]{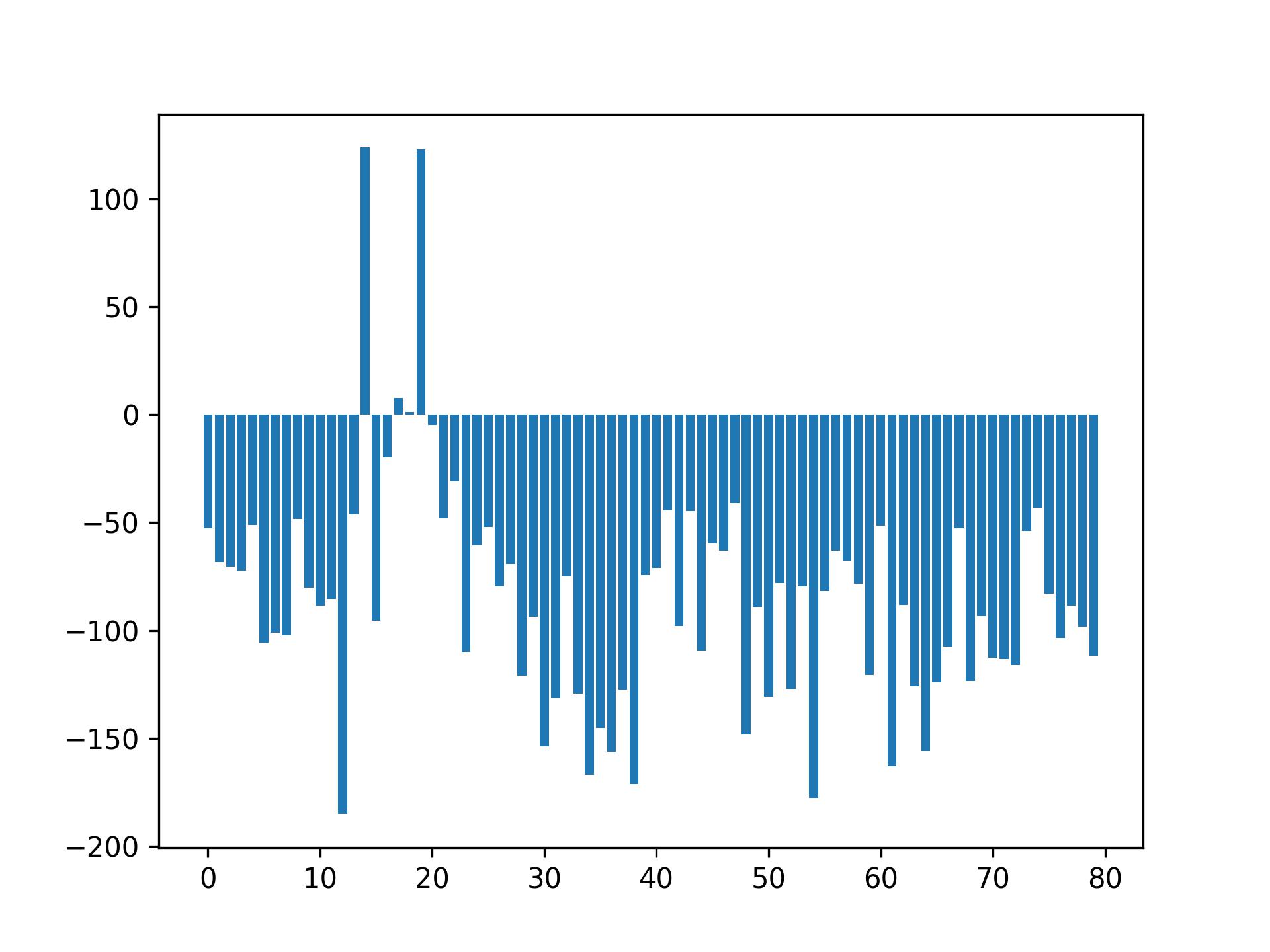} 
        \end{subfigure}
        \hfill
        \begin{subfigure}[b]{0.20\textwidth}  
            \centering 
            \includegraphics[width=\textwidth]{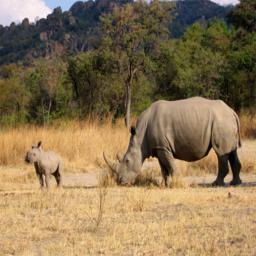}
        \end{subfigure}
        \hfill
        \begin{subfigure}[b]{0.28\textwidth}  
            \centering 
            \includegraphics[width=\textwidth]{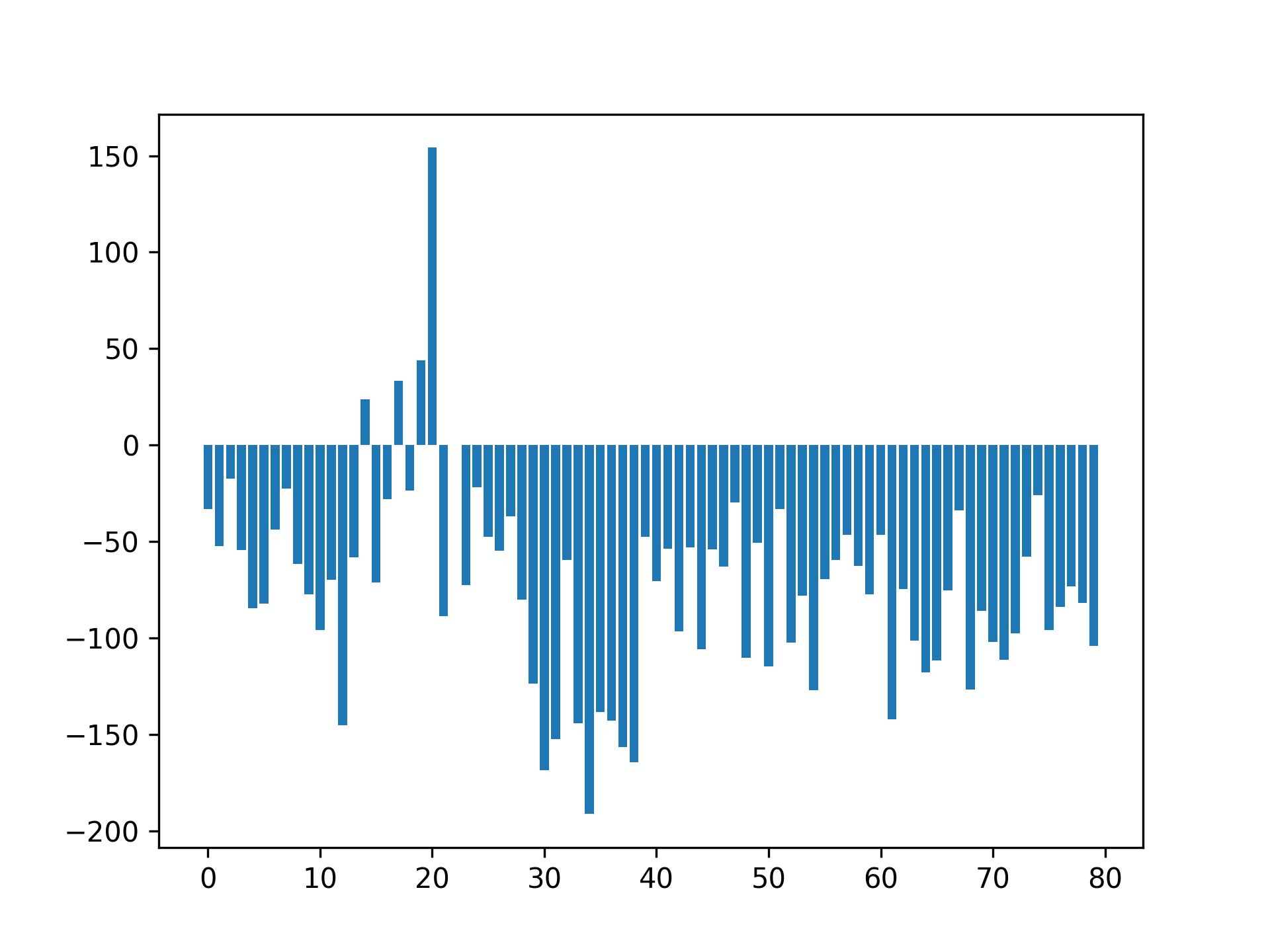} 
        \end{subfigure}
        \hfill
        \begin{subfigure}[b]{0.20\textwidth}  
            \centering 
            \includegraphics[width=\textwidth]{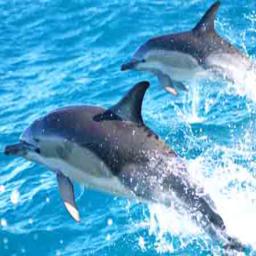} 
        \end{subfigure}
        \hfill
        \begin{subfigure}[b]{0.28\textwidth}  
            \centering 
            \includegraphics[width=\textwidth]{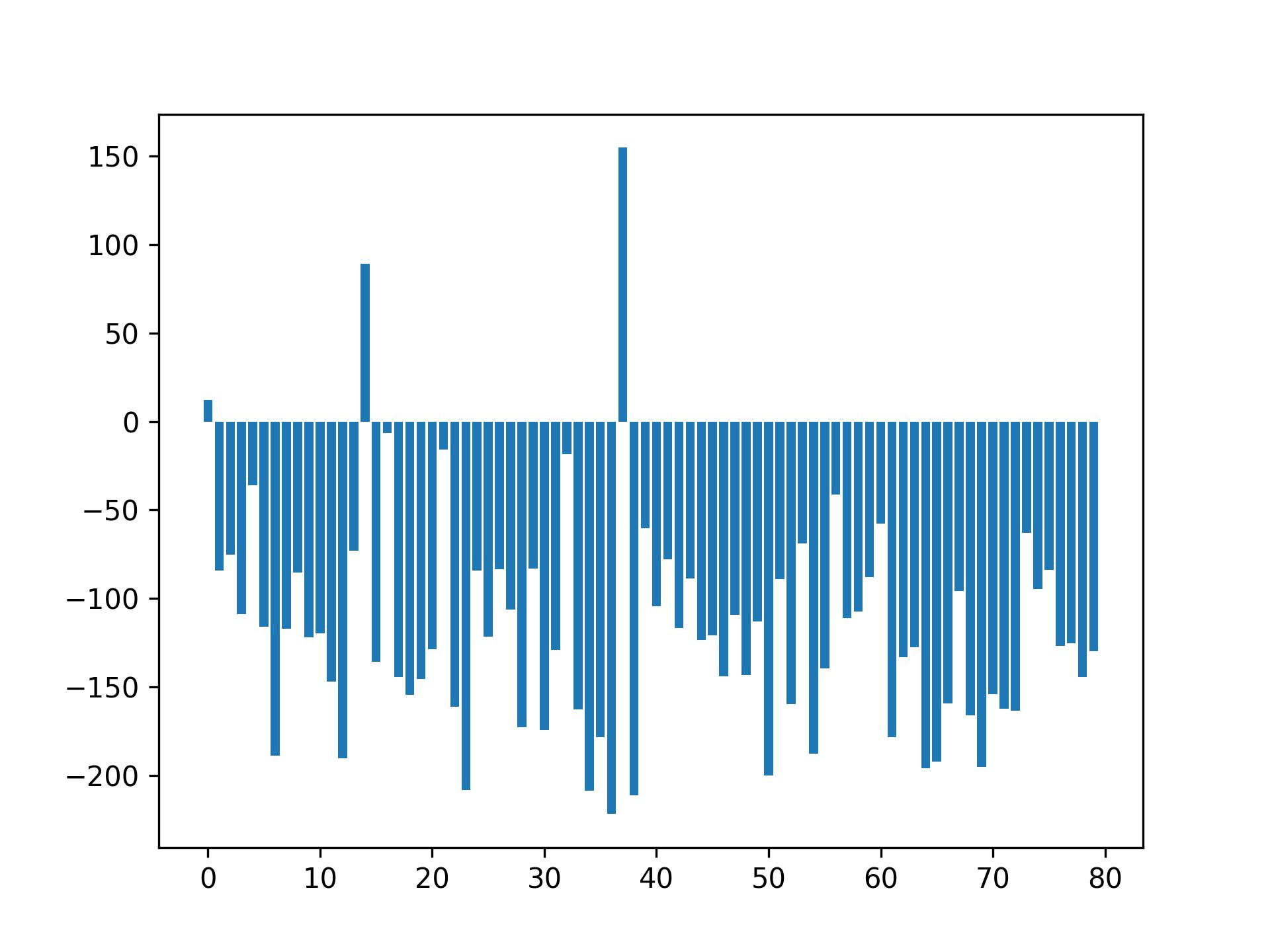}   
        \end{subfigure}
        \hfill
        \begin{subfigure}[b]{0.20\textwidth}  
            \centering 
            \includegraphics[width=\textwidth]{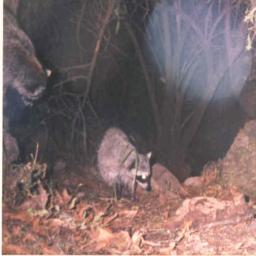}  
        \end{subfigure}
        \hfill
        \begin{subfigure}[b]{0.28\textwidth}  
            \centering 
            \includegraphics[width=\textwidth]{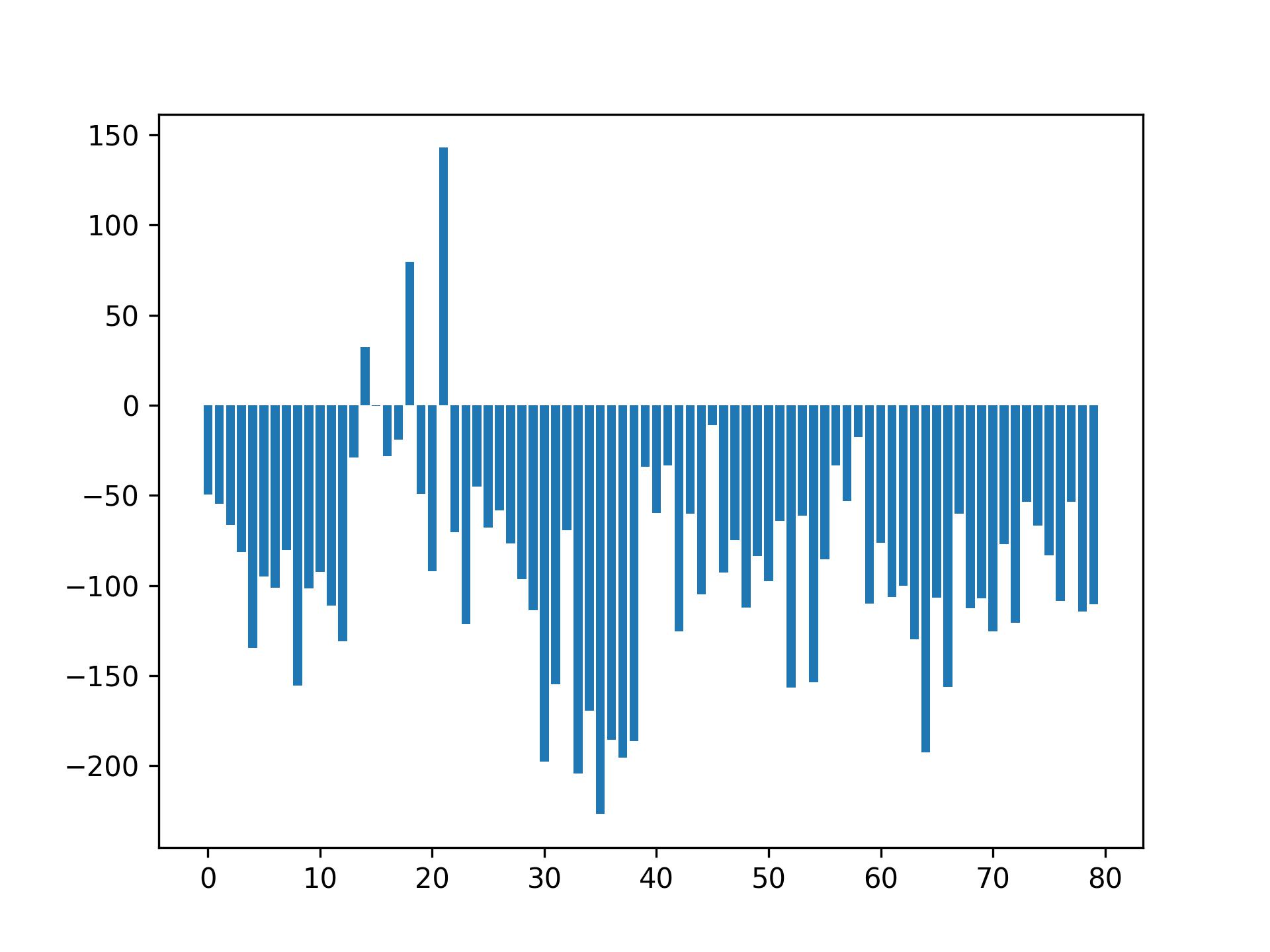}  
        \end{subfigure}
        \caption[  ]
        {\small Examples of OOD detection failures using JointEnergy on MS-COCO/ImageNet-1K datasets. Barplots represent the logit output. } 
        \label{fig:failures}
    \end{figure*}

\begin{figure*}[!t]
        \begin{center}
        \begin{subfigure}[b]{0.24\textwidth}  
            \centering 
            \includegraphics[width=\textwidth]{Images/Fig_5/image_48611_OOD.jpg}
        \end{subfigure}
        \begin{subfigure}[b]{0.32\textwidth}  
            \centering 
            \includegraphics[width=\textwidth]{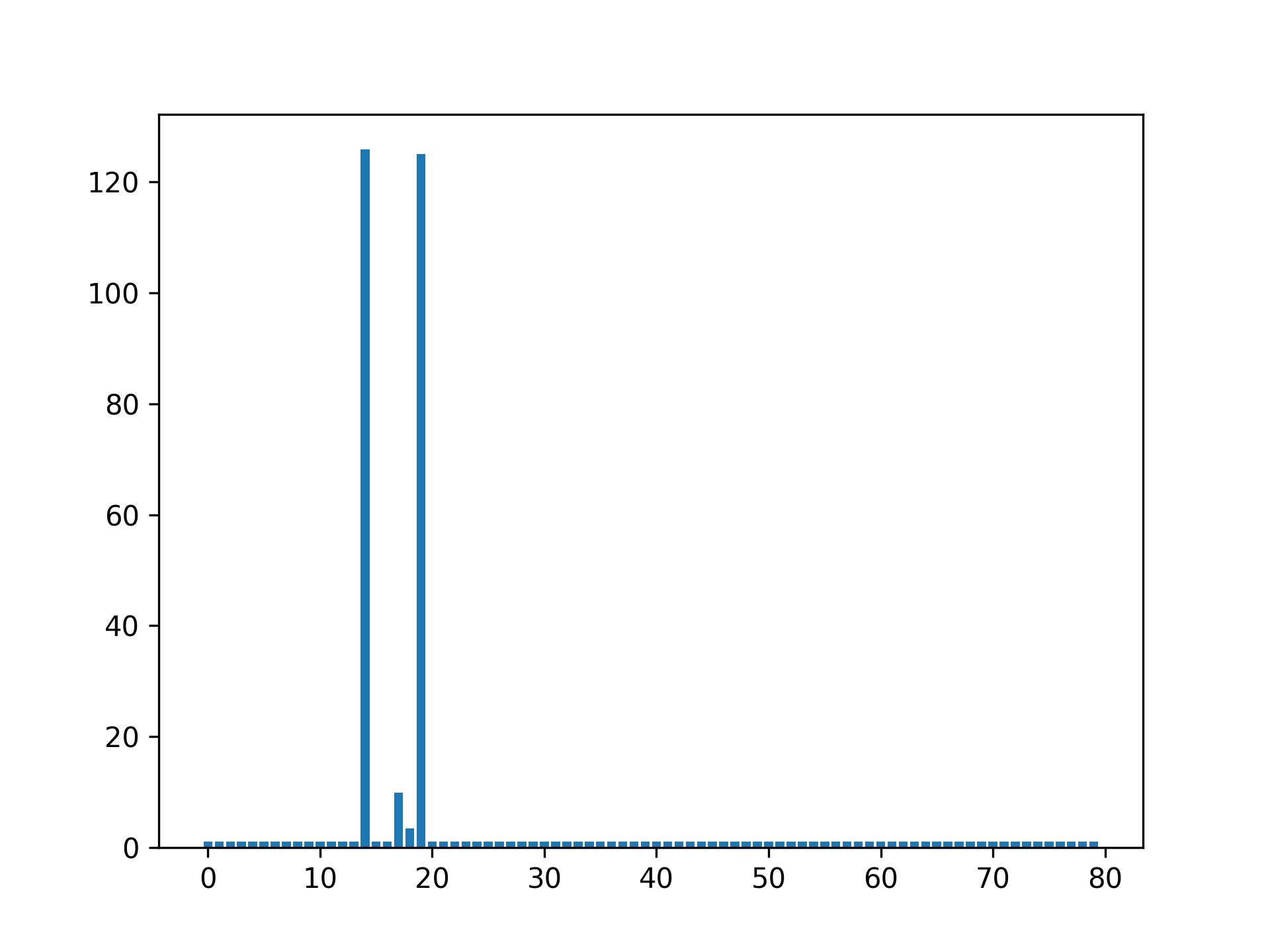} 
        \end{subfigure}
        \begin{subfigure}[b]{0.32\textwidth}  
            \centering 
            \includegraphics[width=\textwidth]{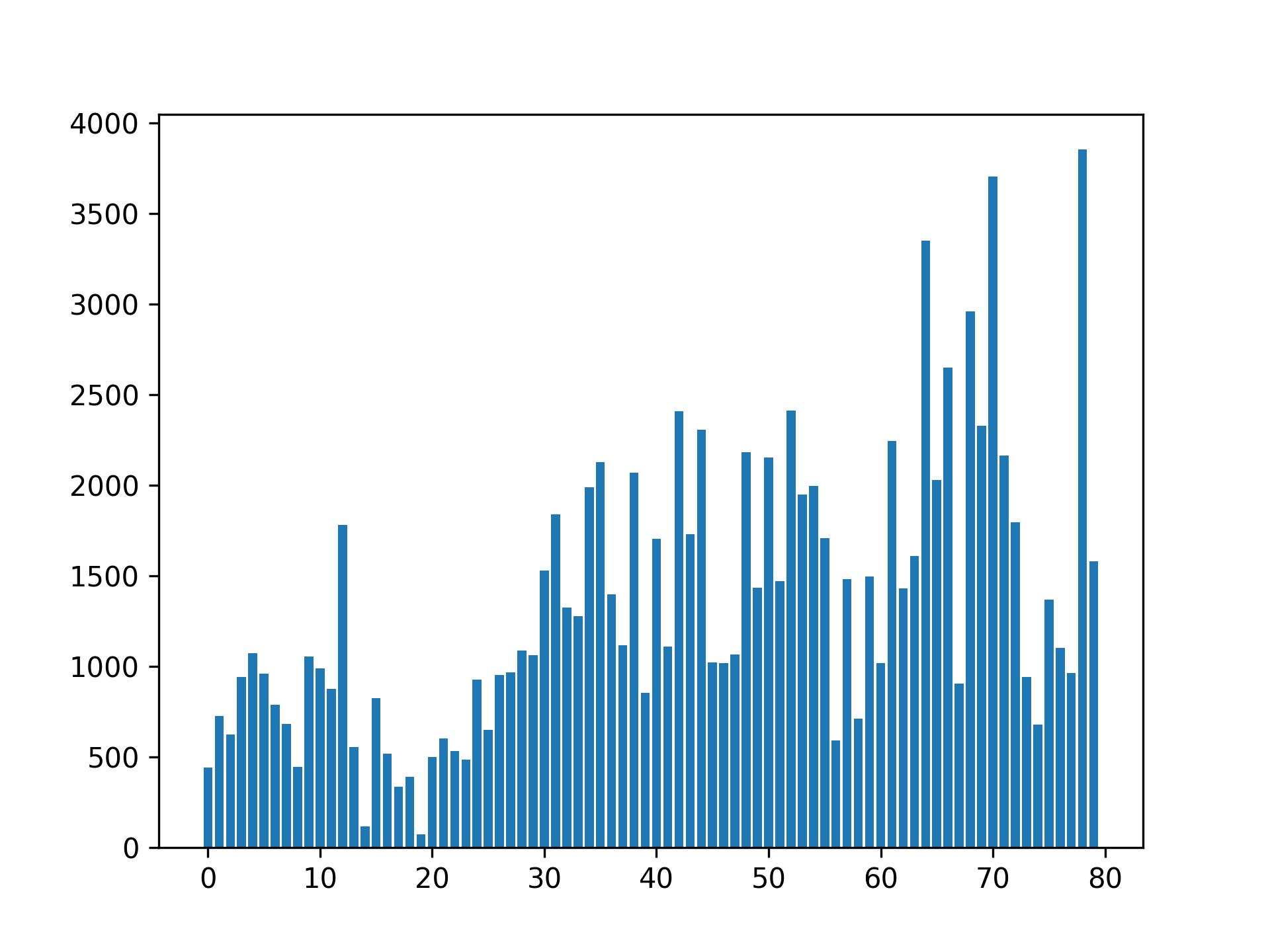}  
        \end{subfigure}
        \\
        \begin{subfigure}[b]{0.24\textwidth}  
            \centering 
            \includegraphics[width=\textwidth]{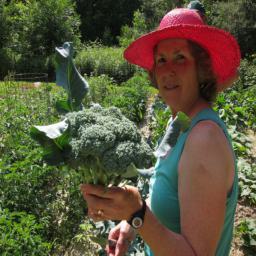}  
        \end{subfigure}
        \begin{subfigure}[b]{0.32\textwidth}  
            \centering 
            \includegraphics[width=\textwidth]{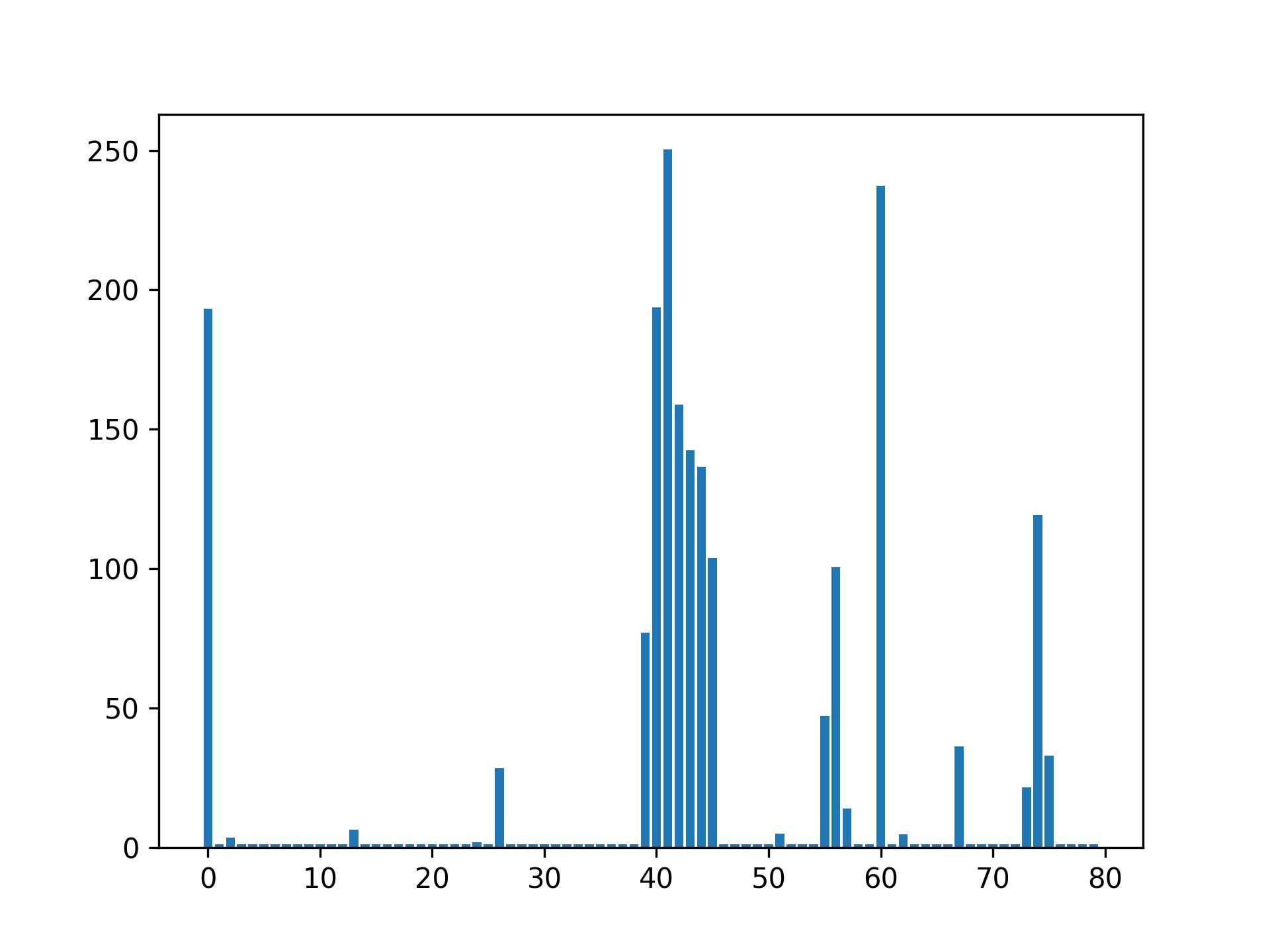}
        \end{subfigure}
        \begin{subfigure}[b]{0.32\textwidth}  
            \centering 
            \includegraphics[width=\textwidth]{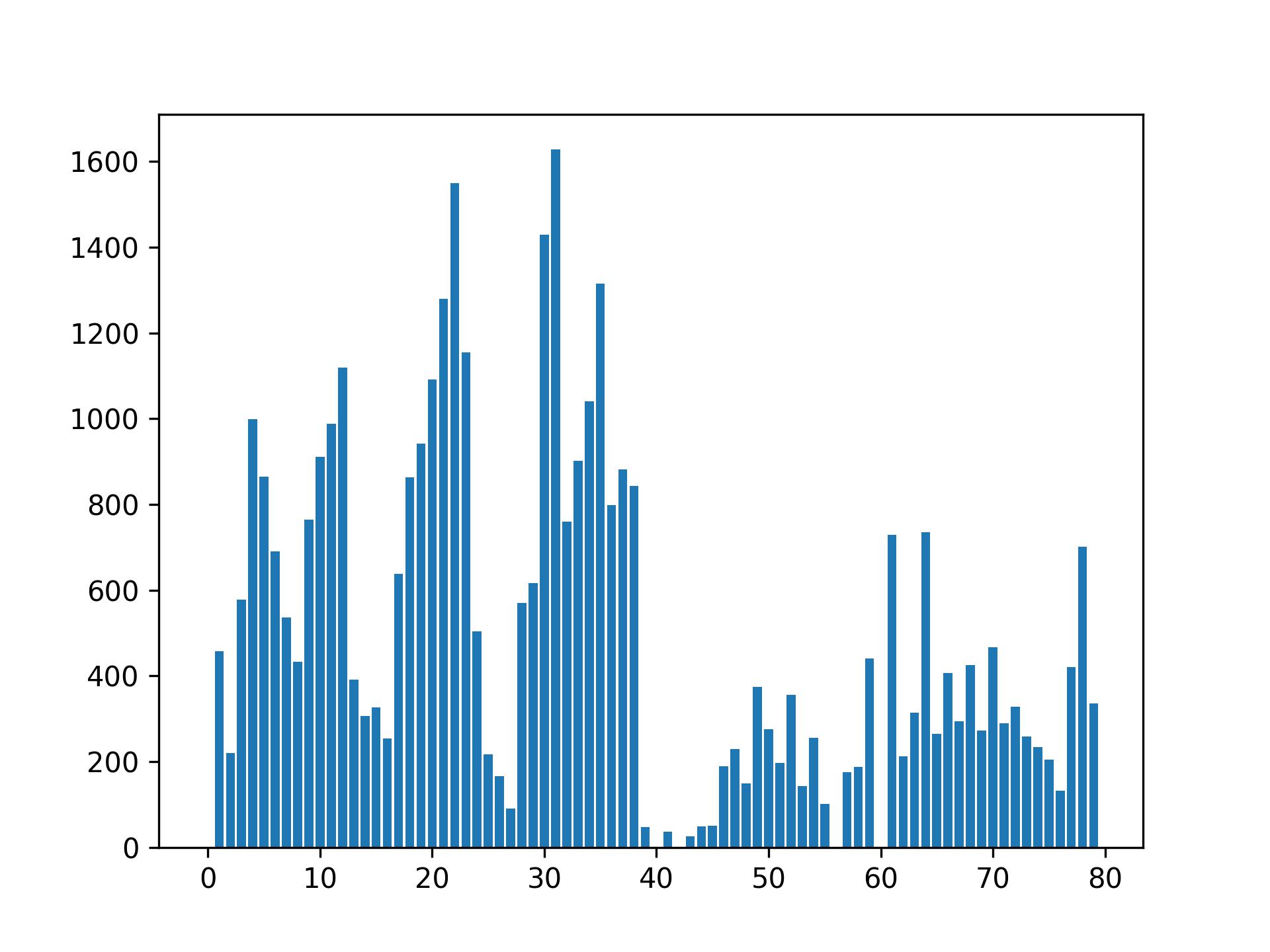}   
        \end{subfigure}
        \end{center}
        \caption[  ]
        {\small Example of successful detection of OOD (top) and In-Distribution (IND) (bottom) with the proposed method on MS-COCO/ImageNet-1K datasets. From left to right: target image, barplot of positive and negative evidence. } 
        \label{fig:success}
    \end{figure*}

Finally, in Fig. \ref{fig:failures}, we present a sample of OOD data that were incorrectly detected as IND data by JointEnergy but were correctly detected by our approach. Additionally, Fig. \ref{fig:success} showcases examples of both OOD and IND data that were successfully detected using our score. As seen in Fig. \ref{fig:failures}, most of the samples have triggered activations in more than one label even though these images are single-labeled data. Although ImageNet and MS-COCO contain different object categories, certain visual similarities, shared by objects (e.g. animals) or backgrounds (e.g. sky, grass, and sea), could potentially lead to uncertain prediction and failure of the OOD detector. However, our proposal is able to prevent such failures by managing both positive and negative evidence. As seen in Fig. \ref{fig:success} (first row), while there is positive evidence of four classes, the approach also presents negative evidence for the same classes. This dual evidence could represent an indicator that the prediction may have been made for unknown categories. In contrast to the OOD data, as seen in the second row of Fig. \ref{fig:success}), in IND data the estimated negative evidence tends to be 0 for prediction with high positive evidence.

\section{Conclusions}
In this paper, we introduced a novel approach for OOD detection in a multi-label classification setting. Our approach exploits the uncertainty predicted by an evidential deep learning framework. We proposed a new Beta-ENN architecture that computes the likelihood and the uncertainty of the input samples. We also defined two new metrics for OOD detection, using positive and negative evidence. Extensive experiments carried out on three public datasets for OOD detection in a multi-label setting showed that our approach outperforms the existing SoTA. Future work will be devoted to extending the current approach in a continual learning scenario, where the training data is presented in a sequential manner.

%
%

\section*{Acknowledgements}
This work is funded by the Government of Chile through its ANID (No. FONDECYT INICIACIÓN 11230262), Beatriu de Pinós Programme and the Ministry of Research and Universities of the Government of Catalonia (2022 BP 00257), 
the Horizon EU project MUSAE (No. \museNo), 2021-SGR-01094 (AGAUR), Icrea Academia'2022 (Generalitat de Catalunya), Robo STEAM (2022-1-BG01-KA220-VET-000089434, Erasmus+ EU), DeepSense (ACE053/22/000029, ACCIÓ), DeepFoodVol (AEI-MICINN, \DFVolNo), PID2022-141566NB-I00 (AEI-MICINN), CERCA Programme / Generalitat de Catalunya, and grants TED2021-132513B-I00 and PID2022-143257NB-I00 (MCIN-AEI). 

\bibliographystyle{splncs04}
\bibliography{main}

\begin{thebibliography}{10}
\providecommand{\url}[1]{\texttt{#1}}
\providecommand{\urlprefix}{URL }
\providecommand{\doi}[1]{https://doi.org/#1}

\bibitem{abati2019cvpr}
Abati, D., Porrello, A., Calderara, S., Cucchiara, R.: Latent space autoregression for novelty detection. In: Proceedings of the IEEE Conference on Computer Vision and Pattern Recognition. pp. 481--490 (2019)

\bibitem{abdar2021uncertaintysurvey}
Abdar, M., Pourpanah, F., Hussain, S., Rezazadegan, D., Liu, L.e.a.: A review of uncertainty quantification in deep learning: Techniques, applications and challenges. Information Fusion  \textbf{76},  243--297 (2021)

\bibitem{cohen2021iccv}
Ben-Cohen, A., Zamir, N., Ben~Baruch, E., Friedman, I., Zelnik-Manor, L.: Semantic diversity learning for zero-shot multi-label classification. In: Proceedings of the IEEE International Conference on Computer Vision. pp. 640--650 (2021)

\bibitem{blundell2015weight}
Blundell, C., Cornebise, J., Kavukcuoglu, K., Wierstra, D.: Weight uncertainty in neural network. In: International conference on machine learning. pp. 1613--1622 (2015)

\bibitem{breunig2000icmd}
Breunig, M.M., Kriegel, H.P., Ng, R.T., Sander, J.: Lof: identifying density-based local outliers. In: Proceedings of the ACM SIGMOD international conference on Management of Data. pp. 93--104 (2000)

\bibitem{chen2019tip}
Chen, L., Zhan, W., Tian, W., He, Y., Zou, Q.: Deep integration: A multi-label architecture for road scene recognition. IEEE Transactions on Image Processing  \textbf{28}(10),  4883--–4898 (2019)

\bibitem{chua2009civr}
Chua, T.S., Tang, J., Hong, R., Li, H., Luo, Z., Zheng, Y.T.: Nus-wide: A real-world web image database from national university of singapore. In: Proceedings of the ACM Conference on Image and Video Retrieval (2009)

\bibitem{deng2009cvpr}
Deng, J., Dong, W., Socher, R., Li, L.J., Li, K., Li, F.F.: Imagenet: A largescale hierarchical image database. In: Proceedings of the IEEE Conference on Computer Vision and Pattern Recognition. pp. 248--255 (2009)

\bibitem{denouden2018arxiv}
Denouden, T., Salay, R., Czarnecki, K., Abdelzad, V., Phan, B., Vernekar, S.: Improving reconstruction autoencoder out-of-distribution detection with mahalanobis distance. arXiv preprint arXiv:1812.02765v1  (2018)

\bibitem{devries2018OOD}
DeVries, T., Taylor, G.W.: Learning confidence for out-of-distribution detection in neural networks. arXiv preprint arXiv:1802.04865v1  (2018)

\bibitem{du2023neurips}
Du, X., Sun, Y., Zhu, J., Li, Y.: Dream the impossible: Outlier imagination with diffusion models. Advances in neural information processing systems  \textbf{37} (2023)

\bibitem{du2022iclr}
Du, X., Wang, Z., Cai, M., Li, Y.: Vos: Learning what you don't know by virtual outlier synthesis. In: International Conference on Representation Learning (2022)

\bibitem{everingham2015ijcv}
Everingham, M., Eslami, S.M.A., Van~Gool, L., Williams, C.K.I., Winn, J., Zisserman, A.: The pascal visual object classes challenge: A retrospective. International Journal of Computer Vision  \textbf{111}(1),  98--136 (2015)

\bibitem{gawlikowski2021survey}
Gawlikowski, J., Tassi, C.R.N., Ali, M., Lee, J., Humt, M., Feng, J., Kruspe, A., Triebel, R., Jung, P., Roscher, R., et~al.: A survey of uncertainty in deep neural networks. arXiv preprint arXiv:2107.03342  (2021)

\bibitem{gawlikowski2023survey}
Gawlikowski, J., Tassi, C.R.N., Ali, M., Lee, J., Humt, M., Feng, J., Kruspe, A., Triebel, R., Jung, P., Roscher, R., et~al.: A survey of uncertainty in deep neural networks. Artificial Intelligence Review  \textbf{56}(Suppl 1),  1513--1589 (2023)

\bibitem{he2022eccv}
He, H., Yuan, Y., Yue, X., Hu, H.: Rankseg: Adaptive pixel classification with image category ranking for segmentation. In: Proceedings of the European Conference on COmputer Vision (2022)

\bibitem{he2016deep}
He, K., Zhang, X., Ren, S., Sun, J.: Deep residual learning for image recognition. In: Proceedings of the IEEE conference on computer vision and pattern recognition. pp. 770--778 (2016)

\bibitem{hendrycks2022icml}
Hendrycks, D., Basart, S., Mazeika, M., Zou, A., Kwon, J., Mostajabi, M., Steinhardt, J., Song, D.: Scaling out-of-distribution detection for real-world settings. In: International Conference on Machine Learning. pp. 8759--8773 (2022)

\bibitem{hendrycks2016baseline}
Hendrycks, D., Gimpel, K.: A baseline for detecting misclassified and out-of-distribution examples in neural networks. In: International Conference on Learning Representations (2016)

\bibitem{hendrycks2017iclr}
Hendrycks, D., Gimpel, K.: A baseline for detecting misclassified and out-of-distribution examples in neural networks. In: International Conference on Representation Learning (2017)

\bibitem{jsang2018subjective}
Jsang, A.: Subjective Logic: A formalism for reasoning under uncertainty. Springer Publishing Company, Incorporated (2018)

\bibitem{kendall2017uncertainties}
Kendall, A., Gal, Y.: What uncertainties do we need in bayesian deep learning for computer vision? Advances in neural information processing systems  \textbf{30} (2017)

\bibitem{lakshminarayanan2017simple}
Lakshminarayanan, B., Pritzel, A., Blundell, C.: Simple and scalable predictive uncertainty estimation using deep ensembles. Advances in neural information processing systems  \textbf{30} (2017)

\bibitem{lee2018neurips}
Lee, K., Lee, K., Lee, H., Shin, J.: A simple unified framework for detecting out-of-distribution samples and adversarial attacks. Advances in neural information processing systems  \textbf{32} (2018)

\bibitem{li2023cvpr}
Li, J.L., Chen, P., Yu, S., He, Z., Liu, S., Jia, J.: Rethinking out-of-distribution (ood) detection: Masked image modeling is all you need. In: Proceedings of the IEEE Conference on Computer Vision and Pattern Recognition. pp. 11578--11589 (2023)

\bibitem{li2023icmr}
Li, Y., Guan, C., Gao, J.: Tsp-tran: Two-stage pure transformer for multi-label image retrieval. In: Proceedings of ACM International Conference on Multimedia Retrieval. pp. 425--433 (2023)

\bibitem{liang2018enhancing}
Liang, S., Li, Y., Srikant, R.: Enhancing the reliability of out-of-distribution image detection in neural networks. In: International Conference on Learning Representations (2018)

\bibitem{lin2014eccv}
Lin, T.Y., Maire, M., Belongie, S., Hays, J., Perona, P., Ramanan, D., Dollar, P., Zitnick, C.L.: Microsoft coco: Common objects in context. In: Proceedings of the European Conference on Computer Vision. pp. 740--755 (2014)

\bibitem{liu2008icdm}
Liu, F.T., Ting, K.M., Zhou, Z.H.: Isolation forest. In: Proceedings of the IEEE International Conference on Data Mining. pp. 413--422 (2008)

\bibitem{maddox2019bayesian}
Maddox, W.J., Garipov, T., Izmailov, P., Vetrov, D., Wilson, A.G.: A simple baseline for bayesian uncertainty in deep learning. In: Advances in neural information processing systems (2019)

\bibitem{rahaman2021uncertainty}
Rahaman, R., et~al.: Uncertainty quantification and deep ensembles. Advances in Neural Information Processing Systems  \textbf{34},  20063--20075 (2021)

\bibitem{ren2021arxiv}
Ren, J., Fort, S., Liu, J., Roy, A.G., Phady, S., Lakshminarayanan, B.: A simple fix to mahalanobis distance for improving near-ood detection. arXiv preprint arXiv:2106.09022v1  (2021)

\bibitem{sensoy2018evidential}
Sensoy, M., Kaplan, L., Kandemir, M.: Evidential deep learning to quantify classification uncertainty. Advances in neural information processing systems  \textbf{31} (2018)

\bibitem{van2020uncertainty}
Van~Amersfoort, J., Smith, L., Teh, Y.W., Gal, Y.: Uncertainty estimation using a single deep deterministic neural network. In: International conference on machine learning. pp. 9690--9700. PMLR (2020)

\bibitem{wang2019aleatoric}
Wang, G., Li, W., Aertsen, M., Deprest, J., Ourselin, S., Vercauteren, T.: Aleatoric uncertainty estimation with test-time augmentation for medical image segmentation with convolutional neural networks. Neurocomputing  \textbf{338},  34--45 (2019)

\bibitem{wang2022cvpr}
Wang, H., Li, Z., Feng, L., Zhang, W.: Vim: Out-of-distribution with virtual-logit matching. In: Proceedings of the IEEE Conference on Computer Vision and Pattern Recognition. pp. 4921--4930 (2022)

\bibitem{wang2021neurips}
Wang, H., Liu, W., Bocchieri, A., Li, Y.: Can multi-label classification networks know what they don’t know? Advances in Neural Information Processing Systems  \textbf{34},  29074--29087 (2021)

\bibitem{wang2022ivc}
Wang, L., Huang, S., Huangfu, L., Liu, B., Zhang, X.: Multi-label out-of-distribution detection via exploiting sparsity and co-occurrence of labels. Image and Vision Computing  \textbf{126},  104548 (2022)

\bibitem{wang2021iccv}
Wang, Y., Li, B., Che, T., Zhou, K., Liu, Z., Li, D.: Energy-based open-world uncertainty modeling for confidence calibration. In: Proceedings of IEEE International Conference on Computer Vision (2021)

\bibitem{wei2022icml}
Wei, H., Xie, R., Cheng, H., Feng, L., An, B., Li, Y.: Mitigating neural network overconfidence with logit normalization. In: International Conference on Machine Learning (2022)

\bibitem{xiao2020neurips}
Xiao, Z., Yan, Q., Amit, Y.: Likelihood regret: An out-of-distribution detection score for variational auto-encoder. Advances in neural information processing systems  \textbf{34} (2020)

\bibitem{yang2022benchmarkOOD}
Yang, J., Wang, P., Zou, D.e.a.: Openood: Benchmarking generalized out-of-distribution detection. In: Advances in Neural Information Processing Systems (2022)

\bibitem{yang2024surveyOOD}
Yang, J., Zhou, K., Li, Y., Liu, Z.: Generalized out-of-distribution detection: A survey. International Journal of Computer Vision  (2024). \doi{0.1007/s11263-024-02117-4}

\bibitem{yang2022eccv}
Yang, Y., Gao, R., Xu, Q.: Out-of-distribution detection with semantic mismatch under masking. In: Proceedings of the European Conference on COmputer Vision (2022)

\bibitem{zhao2023open}
Zhao, C., Du, D., Hoogs, A., Funk, C.: Open set action recognition via multi-label evidential learning. In: Proceedings of the IEEE/CVF Conference on Computer Vision and Pattern Recognition. pp. 22982--22991 (2023)

\bibitem{zhao2023towards}
Zhao, C., Hu, C., Shao, H., Wang, Z., Wang, Y.: Towards trustworthy multi-label sewer defect classification via evidential deep learning. In: ICASSP 2023-2023 IEEE International Conference on Acoustics, Speech and Signal Processing (ICASSP). pp.~1--5. IEEE (2023)

\bibitem{zhoucvpr2022}
Zhou, Y.: Rethinking reconstruction autoencoder-based out-of-distribution detection. In: Proceedings of the IEEE Conference on Computer Vision and Pattern Recognition. pp. 7379--7387 (2022)

\bibitem{zhu2017learning}
Zhu, F., Li, H., Ouyang, W., Yu, N., Wang, X.: Learning spatial regularization with image-level supervisions for multi-label image classification. In: Proceedings of the IEEE conference on computer vision and pattern recognition. pp. 5513--5522 (2017)

\bibitem{zhu2023iccv}
Zhu, K., Fu, M., Wu, J.: Multi-label self-supervised learning with scene images. In: Proceedings of the IEEE International Conference on Computer Vision. pp. 6694--6703 (2023)

\end{thebibliography}
\end{document}